%% file: 0-main.tex
\documentclass[a4paper,fleqn]{cas-sc}

\usepackage[numbers]{natbib}

\def\tsc#1{\csdef{#1}{\textsc{\lowercase{#1}}\xspace}}
\tsc{WGM}
\tsc{QE}

\newcommand{\longquote}[2]{\begin{quote}“\textit{#1}” -- {#2}\end{quote}}

\usepackage{todonotes}

\newcommand{\textbi}[1]{\textbf{\textit{#1}}}

\usepackage{siunitx}
\usepackage{cleveref}
\crefname{figure}{Figure}{Figures}
\usepackage{caption}
\usepackage{subcaption}
\usepackage{hyperref}
\definecolor{lightgray}{gray}{0.9}
\newcommand{\highlight}[1]{\cellcolor{lightgray}#1}

\DeclareMathOperator*{\woe}{woe}

\DeclareMathOperator*{\argmax}{arg\,max}

\begin{document}
\let\WriteBookmarks\relax
\def\floatpagepagefraction{1}
\def\textpagefraction{.001}

\shorttitle{From Evidence to Decision}

\shortauthors{}  

\title [mode = title]{From Evidence to Decision: Exploring Evaluative AI}



%

\author[unsw]{Thao Le}
\cormark[1]
\ead{thao.le1@unsw.edu.au}

\author[uq-cs]{Tim Miller}
\ead{timothy.miller@uq.edu.au}

\author[unimelb]{Liz Sonenberg}
\ead{l.sonenberg@unimelb.edu.au}

\author[csiro]{Ronal Singh}
\ead{ronal.singh@csiro.au}

\author[uq-else]{H. Peter Soyer}
\ead{p.soyer@uq.edu.au}

\affiliation[unsw]{
  organization={School of Computer Science and Engineering, The University of New South Wales},
  city={Sydney},
  country={Australia}}

\affiliation[uq-cs]{
  organization={School of Electrical Engineering and Computer Science, The University of Queensland},
  city={Brisbane},
  country={Australia}}

\affiliation[unimelb]{
  organization={School of Computing and Information Systems, The University of Melbourne},
  city={Melbourne},
  country={Australia}}

\affiliation[csiro]{
  organization={CSIRO's Data61},
  city={Melbourne},
  country={Australia}}

\affiliation[uq-else]{
  organization={Frazer Institute, The University of Queensland, Dermatology Research Centre},
  city={Brisbane},
  country={Australia}}

\cortext[1]{Corresponding author}

\begin{abstract}
  This paper presents a hypothesis-driven approach to improve AI-supported decision-making that is based on the Evaluative AI paradigm - a conceptual framework that proposes providing users with evidence for or against a given hypothesis. We propose an implementation of Evaluative AI by extending the Weight of Evidence framework, leading to hypothesis-driven models that support both tabular and image data. We demonstrate the application of the new decision-support approach in two domains: housing price prediction and skin cancer diagnosis. The findings show promising results in improving human decisions, as well as providing insights on the strengths and weaknesses of different decision-support approaches.
\end{abstract}



\begin{keywords}
 Evaluative AI \sep XAI \sep Decision Support \sep Evidence \sep Skin Cancer Diagnosis
\end{keywords}

\maketitle

\input{1-introduction.tex}
\input{7-lit-review.tex}
\input{2-methodology.tex}

\input{housing-domain.tex}
\input{skin-domain.tex}
\input{5-discussion.tex}
\input{6-conclusion.tex}

\subsection*{Acknowledgements}
This research was supported by \textit{(a)} the University of Melbourne Research Scholarship (MRS), \textit{(b)} Australian Research Council Discovery Grant DP190103414 and undertaken using the \textit{(c)} LIEF HPC-GPGPU Facility hosted at the University of Melbourne. This Facility was established with the assistance of LIEF Grant LE170100200.

\subsection*{Declaration of Competing Interest}
H. Peter Soyer is a shareholder of MoleMap NZ Limited and e-derm consult GmbH and undertakes regular teledermatological reporting for both companies. H. Peter Soyer is a Medical Consultant for Canfield Scientific Inc and a Medical Advisor for First Derm.

\appendix
\section{Appendix}
Supplementary materials are available \href{https://thaole25.github.io/aij-supp/}{online}.


\bibliographystyle{cas-model2-names}

\bibliography{cas-refs}



\end{document}

%% file: 1-introduction.tex
\section{Introduction}

Research on AI-assisted human decision-making has explored several different explainable
AI (XAI) approaches. A common approach is to provide explanations of the AI model's predictions, which can help users understand why the model made that prediction. This approach is called \textit{recommendation-driven AI}, where the model provides a recommendation and additional information about that recommendation. Human decision-makers can then decide whether they should follow the recommendation or not. However, research has shown that AI recommendations, even when accompanied with explanations, are not always helpful in supporting decision-making~\cite{Bansal21,WangEffects22,ZanaCognitive21}. In recent research, \citet{miller2023explainable} argues that the recommendation-driven approach has two main issues: (1) explaining just the AI recommendation limits the user's control, as few alternatives are offered if the user disagrees with the AI; (2) the recommendation-driven approach does not align with the cognitive processes of human decision-making. Therefore, challenges remain in building more effective AI-assisted systems in the future.

One challenge is that the recommendation-driven approach can cause \textit{algorithm aversion}~\cite{dietvorst2015} where people do not trust the AI; or worse, \textit{over-reliance} on the AI system~\cite{vered2023}. Importantly, research has shown that providing explanations does not always reduce over-reliance on AI recommendations, compared to only providing AI predictions~\cite{Bansal21,ZanaCognitive21}. More troubling is the fact that adding more information (i.e., adding explanations for the AI recommendation) is not always helpful and sometimes can lead to worse performance and overconfidence in the wrong information~\cite{oskamp1965overconfidence,omodei2005more}. For instance, explanations can result in people viewing the model's incorrectness as being correct~\cite{KennyFQK21} and people being deceived by incorrect explanations regardless of their expertise~\cite{morrison2024impact}.

Another challenge is that providing recommendations and explanations does not align well with accepted cognitive processes of human decision-making~\cite{peirce1974collected,klein2007data,hoffman2022psychology,miller2023explainable}. Specifically, \citet{hoffman2022psychology} argue that abductive reasoning is an appropriate basis for conceptualising explainable AI, as it involves putting forward hypotheses to explain an event. Further, based on \citet{Yates2012evidence}'s definition of cardinal decision issues, \citet{miller2023explainable} identifies six criteria for good decision aid, including: (1) help to identify options, (2) help to identify possible outcomes for each option, (3) help to judge which outcomes are most likely, (4) help to identify impacts on stakeholders, (5) help to make trade-offs between options, and (6) help to understand the machine's decision. The current recommendation-driven approach (i.e., giving AI recommendations and explanations) only satisfies criteria (6).

To address these issues, \citet{miller2023explainable} proposed a paradigm shift called \textbf{hypothesis-driven} decision support using a conceptual framework termed \textbf{evaluative AI}. The new framework aims to provide evidence for or against a given hypothesis. Rather than telling the decision-makers what to do, the hypothesis-driven approach allows a user to have more control of the decision-making process by incorporating their hypotheses and promoting uncertainty awareness~\cite{le2023dcexplaining,le2024new}. 

In recent work~\cite{le2024new}, we proposed a particular \textit{evidence-informed hypothesis-driven decision-making model} obtained by applying the Weight of Evidence (WoE) framework~\cite{Melis21}. The results showed that the hypothesis-driven approach can improve decision accuracy and reduce over-reliance on AI, but with a small increase in under-reliance when compared to the recommendation-driven approach, in a user study of housing price prediction.
This paper is an extension of that prior work~\cite{le2024new} in two ways. First, we extend the Weight of Evidence (WoE) framework from tabular data to (unstructured) image data by applying concept-based explanations~\cite{zhang2021improving,yuksekgonul2023posthoc} to extract \textit{high-level concepts}, which represent the features of the image. The original WoE framework had been implemented only for tabular data, while extracting features from images is more complex than processing pre-defined features in tabular data. Second, using a case study of supporting skin cancer diagnosis, we report on a human behavioural experiment that investigated the differences between the two approaches (recommendation-driven and hypothesis-driven). To the best of our knowledge, this is the first work that proposes a general tool for \textit{Evaluative AI}.

Our contributions are as follows:
\begin{itemize}
    \item A hypothesis-driven approach for decision support by providing evidence for or against a given hypothesis, built on the Weight of Evidence (WoE) framework.
    \item Extension of the Weight of Evidence (WoE) framework to support both tabular and image data. The latter refers to a \textit{Visual Evaluative AI} model. We also provide public access to this implementation as a Python package~\footnote{\url{https://github.com/thaole25/EvaluativeAI}}.
    \item Evaluating the hypothesis-driven approach compared to the previous decision support models (AI-explanation-only~\cite{GajosIncidental22} and recommendation-driven) through crowdsourcing experiments with laypeople in the housing price prediction domain. The results show that the hypothesis-driven approach can improve decision accuracy and reduce over-reliance on AI, but with a small increase in under-reliance.
    \item Evaluating the hypothesis-driven approach in the skin cancer diagnosis domain. Study participants are individuals with a background in the skin cancer field (e.g., PhD students, postdoctoral researchers, medical doctors and melanographers). The results show that both recommendation-driven and hypothesis-driven have pros and cons, and the hypothesis-driven approach is more preferred by experienced diagnosers. Based on this study, we also propose suggestions for the design of future decision support systems.
\end{itemize}

The following sections of this paper are organised as follows: Section~\ref{sec:lit-review} covers literature review on human cognitive processes when making decisions, different decision-support paradigms and applications of explainable AI in building decision support; Section~\ref{sec:methodology} describes the methodology of the proposed decision support models, including the Weight of Evidence (WoE) model and the concept-based explanation approaches; Section~\ref{sec:housing-domain} presents the study in housing price prediction; Section~\ref{sec:skin-domain} details the study in skin cancer diagnosis; and Section~\ref{sec:discussion} discusses the results of the study and suggestions for future research. The study instruments are available in our \href{https://thaole25.github.io/aij-supp/}{online supplementary materials}.

%% file: 7-lit-review.tex
\section{Background and Related Work}
\label{sec:lit-review}
In this section, we first discuss the cognitive processes in human decision-making and the cognitive biases that form the basis for implementing abductive reasoning in decision support~\cite{hoffman2022psychology,miller2023explainable}. Subsequently, we review different decision support paradigms that inform the experimental designs of our studies. This section concludes with an overview of how explainable AI has been previously applied in decision support, including the reason for choosing the Weight of Evidence (WoE) approach~\cite{Melis21} as the foundation for our work.

    \subsection{Cognitive Processes in Human Decision-Making}
    A well-known psychology theory is the \textit{dual process theory}~\cite{kahneman2011thinking}, which suggests that humans have two different systems for processing information: (1) \textit{System 1} is fast, automatic and intuitive; and (2) \textit{System 2} is slow, deliberate and more accurate. Aligning with this theory, there are two different types of decisions that can be made by humans~\cite{KrawczykAbductive18}: (1) reflexive decision and (2) multi-attribute decision. A reflexive decision is a simple decision made instantly in a short time. It neither involves many attributes in the input nor demands conscious thoughts. By contrast, a multi-attribute decision is a complicated decision that involves many attributes and numerous alternatives.

    We might think that System 2 (slow) is \textit{better} than System 1 (fast) in decision-making and should aim to avoid System 1. However, \citet{miller2023explainable} argues against this view and suggests that we should use the strengths of both systems in designing decision aids. Specifically, the \textit{Naturalistic Decision Making (NDM)} community considers intuition as prior experience that can be used to make rapid and accurate decisions without having to evaluate all options~\cite{Klein15naturalistic,klein2017sources}. The value of intuition can be overlooked if we conduct experiments with laypeople who have no experience in the task and, therefore, fail to capture the advantages of prior knowledge in decision-making~\cite{Klein15naturalistic}.

    Then \textit{\textbf{how should we design decision-support systems that can leverage both System 1 and System 2?}}~\citet{klein2006making1,klein2006making2} examine \textit{sensemaking} from various psychological perspectives, in which \textit{sensemaking} refers to how people make sense of the world. \citet{klein2006making2,klein2007data} present a theory of sensemaking known as \textit{\textbf{Data/Frame Theory}}. The \textit{data} is the information and observations that we use to reconstruct the \textit{frame}, which is a generalisation of a hypothesis. We can question the frame and seek new data to adjust the frame. This process is iterative and done using both System 1 and System 2. People make their decisions first by using their prior knowledge (System 1, System 2) and then search for evidence and make deliberate judgements among plausible options (System 2). Along the same lines, \citet{hoffman2022psychology} argue that Peirce's notion of \textit{\textbf{abductive reasoning}}~\cite{peirce1974collected} best reflects the cognitive processes in the XAI model. Abductive reasoning is the cognitive process when we give hypotheses to explain an occurred event. Based on this foundation, we will comment on the idea of \textit{Evaluative AI}~\cite{miller2023explainable} in Section \ref{subsec:evalAI} and elaborate our method in Section~\ref{sec:methodology}.

    \subsection{Cognitive Biases in Decision-Making}
    Cognitive biases, introduced by \citet{Tversky74}, represent systematic errors in judgment, affecting how individuals perceive input information. These biases can lead to inaccurate and irrational decisions in decision-making contexts. Therefore, understanding the effects of cognitive biases on human-AI decision-making settings, with a particular focus on the application of XAI techniques, is important.

    \citet{Tversky74} outlined several heuristics and biases such as availability, representativeness, and anchoring. The availability heuristic refers to the tendency to rely on the information that is easiest to recall in memory during the evaluation process. Representativeness involves evaluating the probability of an event based on how similar it is to a typical case, rather than measuring the true statistical probability of the event. Anchoring refers to the reliance on the first piece of information encountered when making decisions. \citet{Tversky74}'s foundational work has continued to inform much of the literature on cognitive biases such as confirmation bias~\cite{klayman1995varieties,nickerson1998confirmation}, automation bias~\cite{lee2004trust}, framing~\cite{tversky1981framing}, fixation~\cite{klein2007data}, etc.

    The goal of XAI is to enhance the interpretability and transparency of complex AI models, thereby improving human understanding and trust in AI. Despite the benefits of XAI, human cognitive biases can still influence the decision-making process. For instance, \citet{bertrand2022cognitive} provided a systematic review of the connection between cognitive biases and XAI. They found that cognitive biases can affect or be affected by XAI in various ways, which were classified into four categories: (1) cognitive biases that affect how XAI methods are designed (e.g., explanatory heuristics); (2) cognitive biases that occurred in user studies (e.g., preference for usability over accuracy); (3) cognitive biases that can be mitigated by XAI (e.g., providing prototypes to mitigate representativeness bias); and (4) cognitive biases that can be worsened by XAI (e.g., confirmation bias can lead to over-reliance on the AI). The recommendation-driven approach is an example of how confirmation bias can lead to over-reliance.

    These biases can compromise decision quality even with the help of XAI methods; therefore, de-biasing strategies are needed to reduce the impact of human cognitive biases on decision-making. The idea is to provide explanations for not only the AI's prediction but also other alternative hypotheses, as mentioned in~\cite{bertrand2022cognitive,miller2023explainable}. This approach refers to \textit{hypothesis-driven} decision-making, which can help counter automation bias and fixation by encouraging users to consider evidence of multiple hypotheses. Furthermore, \citet{Rastogi22Fast} proposed a time-based strategy to address anchoring bias by allocating time according to AI confidence.

    \subsection{Decision Support Paradigms}
    In the literature,  there are two workflows that are often used in AI-assisted decision-making: (1) AI-first decision-making; and (2) human-first decision-making. AI-first workflow provides the AI recommendation first and then humans decide if they want to accept or not the recommendation, whereas human-first decision-making requires humans to make a provisional decision before they are provided with any AI recommendations.

        \subsubsection{AI-first Workflow}

        In the AI-first decision-making workflow, it has been demonstrated that participants feel more confident and are also faster in decision-making~\cite{Fogliato22Who}. These participants also rated AI as more practical. In terms of limitations, the anchoring effect is reported to occur more often in the AI-first workflow~\cite{Fogliato22Who,Rastogi22Fast,ZanaCognitive21} in which people overly rely on the AI recommendation (also called \textit{over-reliance}). The anchoring effect~\cite{Tversky74} refers to giving stronger preference to the earlier knowledge rather than doing a full revision and considering the latest evidence. By contrast, \citet{Fogliato21} did not find any significant difference in the participants' performance, which is measured by accuracy between the two workflows (AI-first and human-first). However, they also found that participants are 65\% more likely to revise their answers in the human-first setting than those in the AI-first setting.

        \subsubsection{Human-first Workflow}
        Human-first decision workflow has been shown to help reduce the over-reliance on erroneous AI recommendations~\cite{ZanaCognitive21,Fogliato22Who}. However, experts may interact with decision-making systems differently from laypeople (crowdworkers). For example, \citet{Fogliato22Who,Gaube21} ran studies with radiologists who were the experts. Their task is to review patients' X-ray images. The studies conclude that in human-first workflow with expert participants, they are less likely to leverage AI advice even though the AI is more accurate. In fact, this is referred to \textit{algorithm aversion}~\cite{dietvorst2015} or \textit{under-reliance}.

        A human-first approach called \emph{cognitive forcing}, based on earlier ideas in psychology for interventions that elicit human thinking at decision-making time~\cite{Lambe808}, has been proposed as a way to improve users' engagement and also increase their learning when interacting with the AI \cite{ZanaCognitive21,GajosIncidental22}. Four cognitive forcing designs have been introduced: (1) \textit{On demand}: Participants can only see the AI recommendation when they request it; (2) \textit{Update}: Participants first made a decision without seeing the AI recommendation. Then, they were shown the AI prediction and could update their decision later; (3) \textit{Wait}: Participants had to wait for 30 seconds before the AI decision was shown; and (4) \textit{Only AI explanation}: Providing just the AI explanation and \textit{no AI recommendation}, on the basis that this may help people process the AI explanation more carefully and therefore, improve their knowledge and make better decisions~\cite{GajosIncidental22}.  Importantly, \textit{cognitive forcing} has been shown to reduce \textit{over-reliance} compared to the standard AI suggestion approach, although that study had a limitation in that the AI prediction was always correct. There are also some recognised trade-offs of cognitive forcing designs: more time-consuming~\cite {GajosIncidental22}, and less trust~\cite{ZanaCognitive21}.

        \subsubsection{Evaluative AI (Hypothesis-driven) Paradigm} \label{subsec:evalAI}
        \citet{miller2023explainable} argues that AI-assisted decision support is on the cusp of a paradigm shift. This shift is away from the idea of human-first or AI-first, and into a framework he calls \textbf{evaluative AI}, which is built around the Data/Frame model~\cite{klein2007data}. The key insight of evaluative AI is to not necessarily provide a recommendation, and instead to support the human cognitive decision-making process by providing evidence for or against the particular hypothesis that a human decision-maker is considering. This would help to prevent over- and under-reliance, and would help the decision maker to retain their \emph{internal locus of control}~\cite{Shneiderman16}.

        \subsection{Explainable AI (XAI) in Decision Support}

        In this section, we will review some AI-first explainable AI approaches that have been used to provide explanations for the AI recommendation based on feature analysis. We will also discuss some popular evidence-based explanations that have been used in supporting decision-making.
        
        \subsubsection{Human Decision-Making Reliance on AI Support}
        
        There is no straightforward position regarding when humans are well-calibrated to accept AI-generated advice~\cite{vered2023}. Overall, study participants appear more likely to accept an AI's recommendation when provided with explanations, regardless of the model's correctness~\cite{Bansal21,Jacobs21,WaaNCN21,BussoneClinical15}, whereas less detailed explanations can lead to self-reliance~\cite{BussoneClinical15}. Explanations can increase the accuracy of the human-AI team when the AI is correct, but \textit{decrease} it when it is wrong, resulting in over-reliance on the AI's recommendations. Arguably, this is because the current explanation forms do not provide details of the underlying rationale of the AI model behaviour~\cite{WaaNCN21}. We therefore should be careful when selecting the explanation type as it can have a significant effect on whether users decide to rely on them ~\cite{Caruana15,Lombrozo07}. Moreover, \citet{vasconcelos2023} found that explanations can reduce over-reliance by increasing the task difficulty, easing the explanation difficulty or increasing the benefit of task completion via monetary rewards.

        \subsubsection{Explanations and Cognitive Engagement}

        Besides providing AI recommendations, we now review alternative approaches that can help improve human decision-making. \citet{GajosIncidental22} suggested that providing just the AI explanation and \textit{no AI recommendation} can help people process the AI explanation more carefully and therefore, improve their knowledge and make better decisions. \citet{ChengStrategies19} showed that interactive interfaces can improve users' understanding of how the AI algorithm works and reduce over-reliance on the AI with \textit{cognitive forcing functions}~\cite{ZanaCognitive21}. However, \textit{cognitive forcing functions} can only improve the performance of human+AI teams compared to only explanations given in situations when the AI prediction is incorrect. \citet{naiseh2021nudging} investigated how friction -- small design elements that slow users down -- can improve trust calibration in XAI systems by nudging users to engage more thoughtfully with AI-generated explanations. A study with medical practitioners showed increased interaction with explanations under friction conditions, though it did not significantly deepen cognitive engagement.

        Along the same lines as our work, \citet{cabitza2023let} introduced a \textit{reflective XAI support} that makes users reflect on the reliability of the AI advice by providing evidence both in favour of and against the recommendation. The evidence was presented by showing similar cases, but one where the AI was correct and one where it failed. While the results showed that this approach did not improve diagnostic accuracy, it was appreciated by experienced users for promoting thoughtful evaluation.~\citet{cabitza2024never} then explored a form of explainable AI (XAI) called \textit{pro-hoc explanations}, which present users with similar example cases for each possible outcome instead of offering a single AI prediction. While it led to only modest accuracy improvements, it significantly increased user confidence, especially among less experienced clinicians.~\citet{ma25towards} proposed \textit{Deliberative AI}, built on the Weight of Evidence (WoE) framework~\cite{Melis21}, to provide evidence for conflicting opinions. The results found that Deliberative AI can foster appropriate human reliance on AI and improve decision-making performance.

        Unlike the above approaches, we not only use existing methods but also introduce a novel method. To our knowledge, this is the first work that proposes a general tool for evaluative AI across both tabular and image datasets. We also conducted two user studies -- one with large-scale crowdsourcing and another with a small group of experienced participants -- to empirically evaluate the effectiveness of our approach in improving human decision-making.
        
        \subsubsection{Evidence-based Explanations}
        
        We describe key differences between the Weight of Evidence approach (WoE) and other feature attribution approaches such as LIME~\cite{Ribeiro16} and SHAP~\cite{Lundberg17Unified}.
        
        An approach to generate evidence is the \textit{Weight of Evidence (WoE)} framework~\cite{Melis21}, which can be adapted to meet human-centred design principles. Formally, given a predicted output $y$ for an input $x$, WoE seeks how much \textit{evidence} the input feature $x_i$ gives in favour of (or against) $y$. WoE is similar to feature importance explanations. However, the main difference is that \textit{Weight of Evidence} uses log likelihoods and log odds ratios to generate explanations, whereas LIME~\cite{Ribeiro16} and SHAP~\cite{Lundberg17Unified} find feature importance by modifying the predictive posterior probability in various ways.

        \textbi{We choose the Weight of Evidence approach (WoE) because WoE follows human-centred design principles~\cite{Melis21}}, in which explanations should be \textit{contrastive} (i.e., why the model predicted $y$ instead of alternative $y'$), \textit{exhaustive} (i.e., justify on why every alternative $y'$), \textit{compositional} (i.e., be able to break down into simple components in the prediction), \textit{easily-understandable} (i.e., understandable components) and \textit{parsimonious} (i.e., only provide most relevant facts). \citet{kumar2020problems} argue that SHAP~\cite{Lundberg17Unified} has several human-centred issues, including non-contrastive and non-actionable explanations, and that most people do not have a correct mental model of Shapley values. LIME~\cite{Ribeiro16} does not follow human-centred design principles either. Moreover, LIME~\cite{Ribeiro16} uses a surrogate model to approximate the original one, leading to lower performance and explanations that do not reflect the original model. SHAP~\cite{Lundberg17Unified} can have a high computational cost due to the need to compute Shapley values for each instance. In contrast, WoE~\cite{Melis21} is a probabilistic approach that does not require a surrogate model and can be computed more efficiently.

        A closely related work to~\citet{Melis21} is from~\citet{Poulin06}. \citet{Poulin06} propose a framework called \textit{ExplainD} that uses \textit{additive evidence}. The framework also measures the weight of evidence using a Naive Bayes classifier along with highlighting the negative and positive evidence for a decision. However, the problem being considered is a binary classification. Furthermore, there is still room for improvement by conducting experiments to evaluate the framework. \citet{Kulesza11,Kulesza15} introduce \textit{EluciDebug} in email classification using Multinomial Naive Bayes classifier (MNB). The EluciDebug prototype provides an interface that includes important words and the folder size that both contribute to the email classification. In this example, important words can be referred to as \textit{strength of evidence}. The folder size which describes the number of instances in each class can be referred to \textit{weight of evidence}. Yet the prototype did not specifically give positive and negative evidence in decision-making situations.

        Having more evidence does not always have positive effects on decision-making. \citet{Ratcliff2004} propose stopping rules that can be applied when additional evidence does not change the final decision. More specifically, the \textit{relative stopping rule} refers to when the balance between hypotheses reaches a threshold, meaning evidence for one alternative inherently counts against the other. On the other hand, the \textit{absolute stopping rule} leads to a decision when evidence for a single hypothesis reaches a threshold, with alternatives being considered independently.

%% file: 2-methodology.tex
\section{Methodology}
\label{sec:methodology}

In this section, we present the methodology for our proposed Evaluative AI model, We first introduce the Weight of Evidence (WoE) framework by defining the concept of evidence for both independent and dependent variables. Since tabular data provides pre-defined features, we can directly apply the WoE framework. Then, we introduce the Visual Evaluative AI model, which combines concept-based explanations with the WoE framework, for generating evidence from image data.

    \subsection{Weight of Evidence}
    \label{sec:woe}
    In a classification problem, a hypothesis $h \in Y$, where $Y$ is a set of possible hypotheses. Then, $Y_{-h} = Y\setminus{\{h\}}$ refers to all hypotheses other than $h$. For example, if a doctor asserts a set of hypotheses $Y = \{h_1, h_2, h_3\}$ where \textit{$h_1$ = the patient has Covid}, \textit{$h_2$ = the patient has Influenza} and \textit{$h_3$ = the patient has pneumonia}, then \textit{$Y_{-h_1}$ = the patient does not have Covid} which includes all possible hypotheses except having Covid, that is $Y_{-h_1} = \{h_2, h_3\}$.
    
    We generate the \textit{weight of evidence} for possible hypotheses by applying the Weight of Evidence (WoE) framework, which is a probabilistic approach for analysing variable importance, introduced in the context of explainability by~\citet{Melis21} building on the approach of~\citet{Good1985weight}. It provides a quantitative response to the question of why a model predicted output $h$ for a particular input $X$ in terms of how much each input feature $x_i$ provides in favour of, or against, $h$, relative to alternatives. Through Bayes rule, WoE can be understood as an adjustment to the prior log odds caused by observing the evidence.

    For hypothesis $h$ and input feature $x_i$, weight of evidence, $\woe$,  is defined as follows:
    \begin{equation}
        \woe(h \mid x_i) = \log \frac{P(x_i \mid h)}{P(x_i \mid Y_{-h})} = \log P(x_i \mid h) - \log P(x_i \mid Y_{-h})
        \label{eq:woe}
    \end{equation}
    where $\log P(x_i \mid h)$ is the Gaussian log density for hypothesis $h$.

    Based on the weight of evidence, we say the evidence supports or refutes a hypothesis:
    \begin{itemize}
        \item If $\woe(h \mid x_i) > 0$, evidence $x_i$ supports hypothesis $h$
        \item If $\woe(h \mid x_i) < 0$, evidence $x_i$ refutes hypothesis $h$
        \item If $\woe(h \mid x_i) = 0$, evidence $x_i$ neither supports or refutes hypothesis $h$
    \end{itemize}

    Considering the vector input $X = [x_1, x_2, \dots, x_n]$, the features can be independent or dependent on each other. We can apply Gaussian log density for both independent and dependent variables, depending on how we handle the covariance matrix $\Sigma$. The covariance matrix measures the relationship between two variables, indicating the direction (positive or negative) of how much each pair of features changes together and the strength of their relationship. For independent variables, the covariance matrix is diagonal, while for dependent variables, the covariance matrix is full.

    \paragraph{Independent Variables} For independent variables, the Gaussian log density is based on a univariate Gaussian distribution, which can be simplified as follows:
    \begin{align}
        P(x_i|h) &= \mathcal{N}(x_i; \mu_{i|h}, \sigma_{i|h}^2) = \frac{1}{\sqrt{2\pi \sigma_{i|h}^2}} \exp\left( -\frac{(x_i - \mu_{i|h})^2}{2\sigma_{i|h}^2} \right) \nonumber \\
        \log P(x_i|h) &= -\frac{1}{2} \log(2\pi) -\frac{1}{2} \log(\sigma_{i|h}^2) - \frac{(x_i - \mu_{i|h})^2}{2\sigma_{i|h}^2}
    \end{align}
    where $\mu_{i,h}$ is the mean of $x_i$ for hypothesis $h$, and $\sigma_{i,h}^2$ is the variance of $x_i$ for hypothesis $h$.

    \paragraph{Dependent Variables} For dependent variables, the Gaussian log density is based on a multivariate Gaussian distribution. In this case, we compute the conditional distribution of $x_i$ given the remaining features $X_{-i}$ and the hypothesis $h$. We denote $-i$ as the set of features excluding the feature $i$. We calculate conditional mean, conditional variance and Gaussian log density as follows.

    \begin{align}
        \mu_{i|-i, h} &= \mu_{i|h} + \Sigma_{i, -i|h} \Sigma_{-i, -i|h}^{-1} (X_{-i} - \mu_{-i|h}) \nonumber \\
        \sigma_{i|-i, h}^2 &= \Sigma_{i,i|h} - \Sigma_{i, -i|h} \Sigma_{-i, -i|h}^{-1} \Sigma_{-i, i|h} \nonumber \\
        P(x_i|X_{-i}, h) &= \mathcal{N}(x_i; \mu_{i|-i, h}, \sigma_{i|-i, h}^2) = \frac{1}{\sqrt{2\pi \sigma_{i|-i, h}^2}} \exp\left( -\frac{(x_i - \mu_{i|-i, h})^2}{2\sigma_{i|-i, h}^2} \right) \nonumber \\
        \log P(x_i|X_{-i}, h) &= -\frac{1}{2} \log(2\pi) - \frac{1}{2} \log(\sigma_{i|-i, h}^2) - \frac{(x_i - \mu_{i|-i, h})^2}{2\sigma_{i|-i, h}^2}
    \end{align}

    \paragraph{Subset of features} When we consider a subset of features $S$ for hypothesis $h$, we can calculate the conditional distribution of $X_S$ given the remaining features $X_{-S}$ and the hypothesis $h$. We denote $-S$ as the set of features excluding $S$. We calculate the conditional mean, conditional variance matrix and Gaussian log density as follows.

    \begin{align}
        \mu_{S|-S, h} &= \mu_{S|h} + \Sigma_{S, -S|h} \Sigma_{-S, -S|h}^{-1} (X_{-S|h} - \mu_{-S|h}) \nonumber \\
        \Sigma_{S|-S, h} &= \Sigma_{S,S|h} - \Sigma_{S, -S|h} \Sigma_{-S, -S|h}^{-1} \Sigma_{-S, S|h} \nonumber \\
        \log P(x_S|x_{-S}, y) &= -\frac{|S|}{2} \log(2\pi) - \frac{1}{2} \log |\Sigma_{S|x_{-S}, y}| - \frac{1}{2} (x_S - \mu_{S|x_{-S}, y})^\top \Sigma_{S|x_{-S}, y}^{-1} (x_S - \mu_{S|x_{-S}, y})
    \end{align}

    \paragraph{Mixture model} To compute $P(x_i | Y_{-h})$, we sum over all possible hypotheses $k \in Y_{-h}$:
    \begin{equation}
        P(x_i | Y_{-h}) = \sum_{k \in Y\setminus{\{h\}}} P(x_i | k) P(k)
    \end{equation}

    \subsubsection{How decision-aid models can use WoE to make a decision}
    Using the weight of evidence for each feature $x_i$ as in Equation~\ref{eq:woe}, a decision-aid model can make a prediction based on the total weight of evidence of a hypothesis $h$ by summing up the weight of evidence of this hypothesis based on each feature $x_i$. The total weight of evidence is defined as follows.
    \begin{equation}
        \woe(h) = \sum_{i=1}^n \woe(h \mid x_i)
    \end{equation}
    where $n$ is the number of features.

    The decision-aid model will select the best hypothesis based on the maximum posterior, that is, $y = \argmax_{h \in Y} P(h \mid X)$. If we have the same prior for all hypotheses, we can also use the total weight of evidence as another way to find the best hypothesis using Equation~\ref{eq:woe}. Therefore, a decision-aid model can select the hypothesis with the maximum total weight of evidence as its prediction as follows (only apply to uniform priors).
    \begin{equation}
        y=\argmax_{h \in Y} \woe(h)
        \label{eq:total-woe}
    \end{equation}

    \subsubsection{How WoE can incorporate a human approach to making a decision}

    To assist users with interpretability, \citet{Melis21} complement the display of the magnitude of the weight of the evidence with a notion of the significance level of the evidence, using a scale of seven categories: \emph{decisive-against (\textbf{-}\textbf{-}\textbf{-}), strong-against (\textbf{-}\textbf{-}), substantial-against (\textbf{-}), not-significant (\textbf{N}), substantial-in-favour (\textbf{+}), strong-in-favour (\textbf{+}\textbf{+}), decisive-in-favour (\textbf{+}\textbf{+}\textbf{+})}. The details can be found in the rule-of-thumb guidelines here~\cite{Melisgit}.

    In addition to the weight of evidence of a feature, we suggest it is useful to distinguish the \textit{importance} of a feature -- with importance being domain-specific and determined by the domain expert using the model. Specifically, if a feature has a significant weight of evidence according to the WoE model, but that feature is not seen as important by the human decision-maker, then it is reasonable to anticipate the impact of that evidence on the decision would be reduced by the decision maker. For example, if a clinician looks at a dermatoscopic image and also is aware of some irrelevant but high weight of evidence features such as dense hair,  they should ignore that evidence in making a prediction.

    Formally, by considering the importance of the evidence, we re-define the total weight of evidence from a human decision-making perspective as follows:
    \begin{equation}
        \woe(h) = \sum_{i=1}^n \gamma_i \times \woe(h \mid x_i)
        \label{eq:total_woe}
    \end{equation}

    where $\gamma_i$ is a parameter of feature $x_i$ that adjusts the weight of evidence based on importance, i.e., $\gamma_i > \gamma_j$ represents that feature $x_i$ is more important than feature $x_j$. $\gamma$ can be considered as prior belief of the human decision-maker.

    Then, for the skin cancer example just mentioned, in effect in Equation~\ref{eq:total_woe}, the clinician has set $\gamma = 0$ for that feature.

    \subsection{Visual Evaluative AI}
    \label{sec:veai}

    The Weight of Evidence (WoE) framework in Section~\ref{sec:woe} can be applied to tabular datasets. However, extracting features from images is more complicated than that as it requires sophisticated techniques like convolutional neural networks. Therefore, we extend the WoE by applying concept-based explanations~\cite{yuksekgonul2023posthoc,zhang2021improving} to extract human-understandable concepts from images and put these concepts into the WoE model. Additionally, despite existing methods like Grad-CAM~\cite{selvaraju2017grad} can provide contrastive explanations of various output classes, pixel-level explanations are less effective as they only specify which area of the image is important, but not what is important about that area. \cite{Kim18}.
    
    In this section, we introduce our evidence generation model by combining a concept-based explanation model (e.g., Invertible Concept-based Explanation (ICE)~\cite{zhang2021improving}, Post-hoc Concept Bottleneck Model (PCBM)~\cite{yuksekgonul2023posthoc}) and the Weight of Evidence (WoE) model~\cite{Melis21}. In our experiments, we use \textit{Invertible Concept-based Explanation (ICE)}~\cite{zhang2021improving} as an example of unsupervised concept learning, and \textit{Post-hoc Concept Bottleneck Model (PCBM)}~\cite{yuksekgonul2023posthoc} as an example of supervised concept learning. Combining them together, we propose two models to generate the evidence-based explanations called \textit{ICE+WoE} and \textit{PCBM+WoE}.

    \subsubsection{Concept-based Explanations} We divide concept-based explanations into two categories: (1) supervised learning concepts (concepts are labelled on each image in the training dataset) and (2) unsupervised learning concepts (not having concept labels in the training dataset). Supervised concept learning requires labelled concepts in the training set, or the concepts can be transferred using another labelled dataset~\cite{yuksekgonul2023posthoc}. Unsupervised learning concept methods do not require the concepts to be labelled during the training process. This method is helpful when labelling concepts can be laborious, require expertise, or are not always available. Moreover, unsupervised learning can give users more agency as they can find a new concept that has not been labelled but is still used by a machine learning model. In our application, the evidence will be referred to a concept (or feature) found in the image. Each concept will have a positive/negative quantitative value that shows how much it contributes to the given hypothesis.

    The main difference between ICE+WoE and PCBM+WoE is that the concepts generated by ICE+WoE do not have labels returned by the model. In other words, ICE+WoE identifies only \textit{important concepts} for the classifier but does not assign a name for them. The unlabelled concepts can then be assigned labels by a domain expert. On the other hand, PCBM+WoE provides a concept name for each concept, which is learned from the concept bank. It is important to note that the labelled concepts are not always reliable and still require validation by a domain expert. Furthermore, the number of concepts is fixed based on the concept bank in PCBM+WoE, while the number of concepts can be chosen by the user in ICE+WoE.

    \begin{figure*}[!ht]
        \centering
        \includegraphics[width=\linewidth]{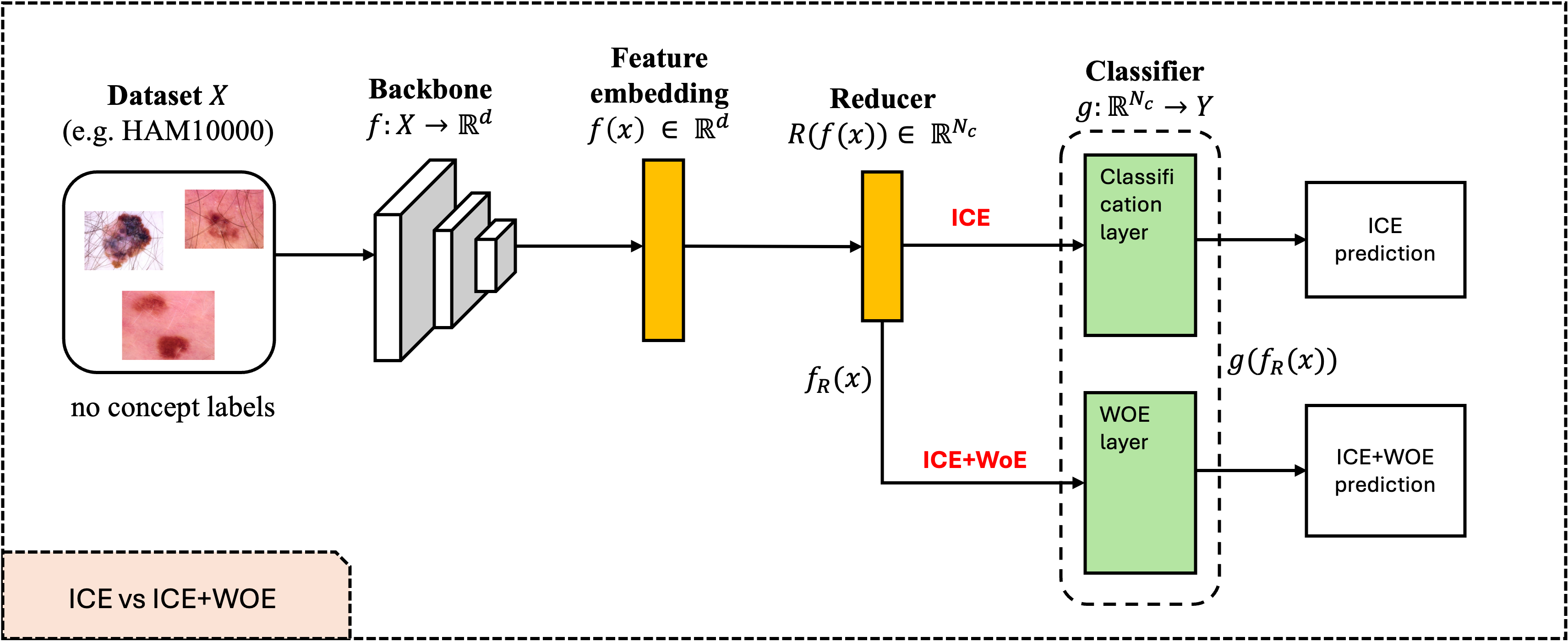}
        \caption{Unsupervised Concept Learning Model (ICE and ICE+WoE)}
        \label{fig:unsupervised-flow}
    \end{figure*}

    \begin{figure*}[!ht]
        \centering
        \includegraphics[width=\linewidth]{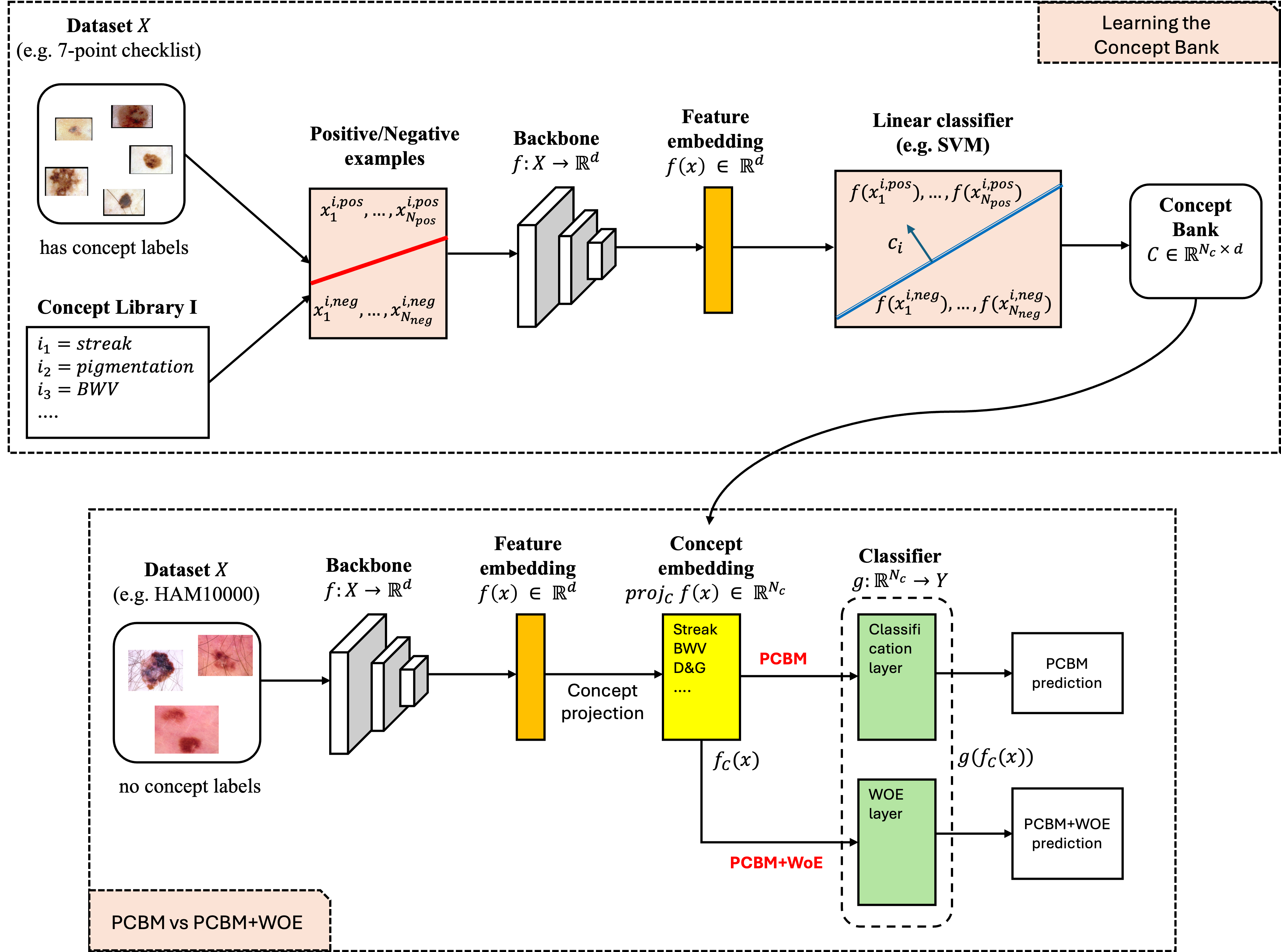}
        \caption{Supervised Concept Learning Model (PCBM and PCBM+WoE)}
        \label{fig:supervised-flow}
    \end{figure*}

    \subsubsection{ICE+WoE and PCBM+WoE}

    We will now explain the implementation of the combined models of concept-based explanations and the Weight of Evidence (WoE) model, depending on the concept learning method (unsupervised or supervised). Figure~\ref{fig:unsupervised-flow} shows an overview of the unsupervised concept learning model, and how it can be combined with Weight of Evidence (WoE)~\cite{Melis21}. Figure~\ref{fig:supervised-flow} shows an overview of the supervised concept learning model using a transferred concept bank, and how it can be combined with Weight of Evidence (WoE)~\cite{Melis21}.

    Formally, let $f: \mathcal{X} \to \mathbb{R}^d$ be the pre-trained backbone model (e.g. ResNet, ResNeXt) where $\mathcal{X}$ is the input space and $d$ is the size of the embedding space. The dimension $d$ of the feature embedding space is often large (e.g., 2048 features). We then aim to reduce the dimensionality by using different techniques, either a reducer (for unsupervised learning) or a concept bank (for supervised learning). Next, we get a set of selected concepts and put them into a classifier layer $g: \mathbb{R}^{N_c} \to Y$ where $Y$ is the set of output classes, $N_c$ is the number of concepts. To implement the combined models ICE+WoE and PCBM+WoE, we replace the classifier layer of ICE and PCBM with the WoE model.

    The following sections will describe in detail unsupervised concept learning and supervised concept learning based on their main difference after the feature embedding layer. For unsupervised concept learning, we use a reducer to reduce the dimensionality of the feature embedding space. The learned concepts may not be related to the concepts used in the domain; for example, they could be invalid dermatological concepts or unknown to domain experts. In contrast, the supervised concept learning technique learns the concept bank and transfers it to create a concept embedding layer. This technique requires labelling concepts on the training images, which is expensive.

    \paragraph{Invertible Concept-based Explanation (ICE)}
    ICE~\cite{zhang2021improving} is unsupervised concept-based explanation approach that applies a reducer $R$ such as Non-negative Matrix Factorization (NMF) to reduce the dimensionality of the feature embedding space. NMF is a matrix factorization method that factorizes the feature matrix into two or more non-negative matrices~\cite{olah2018building}. We can also apply other dimensionality reduction methods such as PCA (Principal Component Analysis), which is a linear transformation method that projects the feature matrix into a lower-dimensional space. Eventually, we get a set of concepts $f_R(x) \in \mathbb{R}^{N_c}$ at the reducer layer.

    \paragraph{Post-hoc Concept Bottleneck Model (PCBM)}
    PCBM~\cite{yuksekgonul2023posthoc} is a supervised concept-based explanation approach that learns the concept bank $\mathcal{C}$ and transfers it to create a concept embedding layer. We learn the concept bank by training a linear SVM to separate \textit{positive concept examples} (contain the concept) and \textit{negative concept examples} (do not contain the concept) in the embeddings based on CAV (Concept Activation Vector) approach~\cite{Kim18}. Importantly, the dataset used to learn the concept bank can be different from the dataset used in the task prediction. Therefore, when the training dataset does not have concept labels, we can annotate concepts by using another dataset that has concept labels and is in the same domain.

    Formally, the concept library is defined as $I = \{i_1, i_2, ..., i_{N_c}\}$. The concept library can be selected by domain experts or learned from the data~\cite{Ghorbani2019concept}. For each concept $i$, there are a set of feature embeddings for positive examples $P_i = \{f(x^{i,pos}_{1}), ..., f(x^{i,pos}_{N_{pos}})\}$ and for negative examples $N_i = \{f(x^{i,neg}_{1}), ..., f(x^{i,neg}_{N_{neg})}\}$ where $N_{pos}$ and $N_{neg}$ are the number of positive and negative examples, respectively. Next, a linear SVM is trained using $P_i$ and $N_i$ to learn CAV (the normal to the SVM's decision boundary) for concept $i$, denoted as $c_i$. Finally, we get the concept matrix $\mathcal{C} \in \mathbb{R}^{N_c \times d}$ at the concept embedding layer.

    Let $g: \mathbb{R}^{N_c} \to Y$ be the classifier. To learn the classifier in PCBM, we minimise the loss function:
    \begin{equation}
        \min_{g}\underset{(x, y)\sim \mathcal{D}}{\mathbb{E}} [\mathcal{L}(g(f_{\mathcal{C}}(x)), y)] + \frac{\lambda}{N_cK}\Omega(g)
    \end{equation}
    where $f_{\mathcal{C}}(x)$ is the projection onto the concept subspace,  $\mathcal{L}(\hat{y}, y)$ is a loss function such as cross-entropy loss, $\Omega(g)$ is a complexity measure to regularize the model, and $\lambda$ is the regularization strength. In PCBM, a linear classifier such as a stochastic gradient descent model is implemented in this layer.

    \paragraph{Weight of Evidence} We replace the original classifier layer in ICE and PCBM with the WoE model. Similar to the WoE model in Section~\ref{sec:woe}, we now calculate the weight of evidence for each concept. For hypothesis $h$ and concept $c_i$, which is equivalent to feature $x_i$ in~\ref{sec:woe}, the weight of evidence $\woe$ is defined as follows.

    \begin{equation}
        \woe(h \mid c_i) = \log \frac{P(c_i \mid h)}{P(c_i \mid Y_{-h})} = \log P(c_i \mid h) - \log P(c_i \mid Y_{-h})
        \label{eq:woe-concept}
    \end{equation}

    In the implementation of Visual Evaluative AI, we use the WoE \textit{without} independence assumption.

%% file: housing-domain.tex
\section{Study 1: Evaluative AI on Tabular Housing Price Data}
\label{sec:housing-domain}

In this section, we will describe the implementation of \textit{Evaluative AI} on tabular data (housing price prediction domain). Experimenting with this task, we compared the \textit{hypothesis-driven} approach with two state-of-the-art decision-making approaches (\textit{recommendation-driven} and \textit{AI-explanation-only}) using quantitative measures for \emph{efficiency}, \emph{performance} and \emph{reliance} and qualitative analysis of \emph{information use}. In the terminology of a recent review of XAI evaluation~\cite{lai2023}, the first two points of comparison are a form of evaluation with respect to the decision task, and the latter two focus on users' perception and use of the AI system itself.

In selecting a decision-making task, we identified requirements similar to those used in other studies of how explanations can assist human decision-makers interacting with AI decision support, e.g.~\citet{vasconcelos2023}: the task should not be too easy for humans to complete without a decision aid, but also, as we were using lay subjects from Prolific for this particular study, the task cannot require specialist knowledge.

We chose a version of the \textit{housing price prediction} task studied previously in an XAI context~\cite{abdul2020cogam,chiang2022}. In this task, participants are provided with information about house features and other information which varies by experimental condition and are asked to choose whether the given house would have a sale price of \textit{low}, \textit{medium} or \textit{high}. As noted by others~\cite{chiang2022}, real estate valuation is a domain where ML models have been developed to help people make better decisions, predicting house prices is a task that lay people may need to do in real life, so it is not unrealistic to expect they have sufficient day-to-day knowledge to make predictions and decide whether or not to rely on an AI model.

    \subsection{Dataset and Model Implementation}

    To build our model, we used the Ames Housing Dataset~\cite{de2011ames} and the open source code on GitHub~\cite{amesgit} for data pre-processing. The data after pre-processing has a total of 2616 instances and 28 features. We processed the dataset further by converting the house price into three output classes (\textit{low price}, \textit{medium price} and \textit{high price}). We also balanced the dataset to ensure that the three classes had the same number of instances by using Near-Miss Undersampling. Finally, we had a total of 1920 instances with 640 instances for each class.

    We selected six features for the human experiment in the house-price decision-making task by applying a Gradient Boosting Classification model over the data. Considering domain-specific decision-making about house prices, we propose there to be three important features (\textit{quality of construction}, \textit{house age} and \textit{location}) and three unimportant features (\textit{fireplaces}, \textit{kitchen quality} and \textit{central air conditioning}). We divided the dataset into 80\% for the training set and 20\% for the test set. 

    \subsection{Experimental Conditions}
    \label{sec:experiment-design}

    All participants\footnote{We received ethics approval from our institution before conducting the human experiment (ID: 23208).} were given the six house feature values plus other information, which varied by condition as set out below.
    Participants then chose whether the given house would have a price of \textit{low}, \textit{medium} or \textit{high}. Using a \textit{between-subject design}, participants were randomly assigned to one of three conditions:
    \begin{itemize}
        \item (C1) \textit{Recommendation-driven}: Participants see the AI prediction (i.e., either \textit{low} or \textit{medium} or \textit{high}) \emph{and} also the weight of evidence for that prediction;
        \item (C2) \textit{AI-explanation-only}: Participants see the weight of evidence associated with the AI prediction, but the AI prediction itself is hidden;
        \item (C3) \textit{Hypothesis-driven}: Participants see the weight of evidence for \emph{all} hypotheses (\textit{low}, \textit{medium} and \textit{high}), but the AI prediction itself is hidden.
    \end{itemize}

    Although participants in the \textit{AI-explanation-only} and \textit{hypothesis-driven} conditions did not see a recommendation, it was expected that the displayed information from the WoE framework would provide insight that participants could use to support their decision-making. We note a similar AI-explanation-only approach has been explored previously~\cite{GajosIncidental22}.

    \subsection{Research Questions and Hypotheses}

    Our overarching research questions were as follows:
    \begin{itemize}
        \item \textbf{RQ1}: \emph{(Efficiency)} What form of AI assistance helps participants make faster decisions?
        \item \textbf{RQ2}: \emph{(Performance)} What form of AI assistance helps participants make better decisions?
        \item \textbf{RQ3}: \emph{(Reliance)} What form of AI assistance helps reduce over-reliance and under-reliance?
        \item \textbf{RQ4}: \emph{(Information use)} How do people make decisions differently in \textit{recommendation-driven}, \textit{AI-explanation-only} and \textit{hypothesis-driven} approach?
    \end{itemize}

    For \textbf{RQ1}, we evaluated the participants' speed in making a decision. We use the most common metric - \textit{completion time} to measure the time taken on the task. The corresponding hypotheses for this question are:
    \begin{itemize}
        \item \textbf{H1a/b}: \textit{(C3) Hypothesis-driven} approach will cost less time to finish the task than \textit{(C1) Recommendation-driven} and \textit{(C2) AI-explanation-only}.
    \end{itemize}

    For \textbf{RQ2}, we evaluated the quality of the decision. In the task, we asked the participants to assign the likelihood for each price range (low/medium/high) where 100 is the most likely and 0 is the least likely. The sum of three likelihoods must be equal to 100. We expect the participants to be confident when they make a correct prediction, and \textit{not be confident when they make a wrong decision}. We apply \textit{Brier score} as explained below to measure the task performance. The hypotheses for this question are:
    \begin{itemize}
        \item \textbf{H2a/b}: \textit{(C3) Hypothesis-driven} approach will help participants make better decisions than \textit{(C1) Recommendation-driven} and \textit{(C2) AI-explanation-only}.
    \end{itemize}

    For \textbf{RQ3}, we investigated the participants' capability of appropriately calibrating their decision. Participants should follow the model's prediction when it is correct and should not use the model's prediction when it is wrong. We applied two measures \textit{over-reliance} and \textit{under-reliance} as shown below with the following hypotheses:
    \begin{itemize}
        \item \textbf{H3a}: \textit{(C3) Hypothesis-driven} can reduce over-reliance compared to \textit{(C1) Recommendation-driven}.
        \item \textbf{H3b}: \textit{(C3) Hypothesis-driven} can reduce under-reliance compared to \textit{(C2) AI-explanation-only}.
    \end{itemize}

    For \textbf{RQ4}, we looked into the text written by participants when they explained why they selected an option after each question to know how they used the provided information in each decision-making approach to make their decisions. Therefore, we can identify the limitations of each approach and the generated evidence that led the participants to make a wrong decision.

    \subsection{Measures}
    We took the following measures:

    \begin{enumerate}
        \item \textit{Task Efficiency} (Completion time): The time participants take to complete each task.
        \item \textit{Task Performance} (Brier score): This metric quantifies the effectiveness of task performance in terms of accurate decision outcomes. The formula is:
            \begin{equation}
                \text{BS}_{p} = \frac{1}{N} \sum_{i=1}^{N} (C_{p,i} - A_{p,i})^2 
            \end{equation}
            where:          
            $C_{p,i}$ is the likelihood level of participant $p$ in question $i$, ranging from 0 to 1 (the likelihood level of a participant refers to how confident they are when answering the question); $A_{p,i}$ is the answer score of participant $p$ in question $i$, either 0 (wrong answer) or 1 (right answer); N is the number of questions for each participant. The best Brier score (i.e. equal to 0) for an individual task is when a participant answers the task correctly and gives it a 100\% likelihood (or alternatively, a wrong answer but with 0\% likelihood). Therefore, a participant has better task performance when they have a lower Brier score. The Brier score measures decision accuracy but awards a higher score for a correct answer when a participant is confident, and a lower penalty for an incorrect answer when not confident. This mitigates problems where participants guess answers (i.e. have low confidence in their answers). Brier score has potential for real-world applications, as it accounts for both the correctness of a decision and the confidence of the decision-maker.
    \end{enumerate}

    We then measure the \textit{over-reliance} and \textit{under-reliance}~\cite{WangEffects22}. Since study participants can only see the AI recommendation in C1 (\textit{Recommendation-driven}), we measure \emph{agreement}: whether participants have the same prediction or differ from the model's prediction in the other two conditions.

    \begin{enumerate}
            \setcounter{enumi}{2}
            \item \textit{Over-reliance}: the fraction of tasks where participants have the same decision as a model's prediction when it was wrong: ${\Sigma_{i} (A_{p,i}=M_i=0})/{\Sigma_i (1-M_i)}$, where $A_{p,i}$ is as above and $M_i=1$ if the model is correct and 0 otherwise.
        
        \item \textit{Under-reliance}: the fraction of tasks where participants have a different decision from a model's prediction when it was correct: ${\Sigma_{i} (A_{p,i}\neq M_i=1)}/{\Sigma_i M_i}$.
    \end{enumerate}

    \subsection{Conduct}
    \label{sec:housing:conduct}

    We conducted \textbf{\textit{two}} separate human experiments in which participants were given the same task in the form of a question set, with the only difference being the way they answered the question. 

    In experiment 1, participants were asked to make a decision about the relative likelihood for each price range (\textit{low}/\textit{medium}/\textit{high}) of given house instances. We answer \textbf{RQ1}, \textbf{RQ2} and \textbf{RQ3} by analysing the results of four measures mentioned above (completion time, Brier score, over-reliance and under-reliance).

    In experiment 2, we recruited a new and smaller cohort and asked them to do the same tasks as in experiment 1, but in addition, we asked participants to explain their decisions using free text. We conduct this experiment separately from the quantitative data in experiment 1 because asking participants to explain their reasoning cognitively forces them to engage with the instance, interfering with their natural decision-making process, and therefore potentially affecting the quantitative results. We then performed a deductive analysis of their explanations to answer \textbf{RQ4}.

    Each experiment was designed as a Qualtrics\footnote{https://www.qualtrics.com} survey and participants accessed the survey through Prolific\footnote{https://www.prolific.com}. The experiment required a maximum of 25 minutes to finish. There were 12 house instances given, equivalent to 12 questions. These 12 questions were evenly distributed into four question categories: (1) where the model gives \textit{correct} predictions with \textit{high uncertainty}, (2) where the model gives \textit{correct} predictions with \textit{low uncertainty}, (3) where the model gives \textit{wrong} predictions with \textit{high uncertainty} and (4) where the model gives \textit{wrong} predictions with \textit{low uncertainty}. Thus, there are three questions in each category. Study participants were \textit{not} informed about in which category the question belongs. The questions are ordered randomly in the experiments.

    The uncertainty is measured by the cross entropy as follows.
    \begin{equation}
        u(h) = -\sum_{h \in H} p(h) \log p(h)
    \end{equation}
    where $u(h)$ is the uncertainty level of hypothesis $h$ given the probabilistic output is $p(h)$. To ensure there was a clear difference between high and low uncertainty, we selected instances with \textit{low uncertainty} by choosing instances with entropy less than 0.3, and \textit{high uncertainty} by choosing instances with entropy greater than 0.7. Participants did not know the certainty nor how many test instances were correct/incorrect. Each participant was paid a minimum of \pounds4 for their time, plus a bonus of \pounds2 if they could answer at least 9 out of 12 questions correctly. Participants were also given a plain language statement, and consent form and did a training phase with 3 example questions before answering the 12 test questions.

    For \textbf{RQ4} the text is a response to ``Can you please explain why you selected this option?". We analysed a total of 12 (questions) $\times$ 95 (participants) = 1140 responses. The final analysis includes 1,031 responses after removing 109 responses due to poor quality.  Each response is assigned to at least one category (or code): \textit{Using feature values} or \textit{Using evidence}. We then perform a simple deductive analysis by reading each response and assigning the relevant codes. We explain each code as follows: (1) \textit{Using Feature Values}: participants rely on the feature values and their background knowledge to make the final decision without using the model evidence; and (2) \textit{Using Evidence}: Participants rely on the evidence provided by the model and possibly their background knowledge to make the final decision.

    We chose these two codes based on the idea of \textit{machine explanation} and \textit{human intuition}~\cite{Chen2022,chen2023understanding}. Specifically, \textit{using evidence} refers to using the machine explanation and therefore, making use of the model evidence to support decision-making. On the other hand, \textit{using feature values} is relevant to using people's intuitions of the task based on the input feature values. Therefore, using the qualitative analysis, we explore how people use the model evidence and their intuitions in the three decision-making approaches.

    \subsection{Participants}
    \paragraph{Experiment 1}
    Using the power analysis for the F-test for one factor ANOVA and assuming the power of 0.8 and significant alpha of 0.05, we found that a sample size of 300 participants in three groups guarantees a small effect size of 0.2. In total, we recruited $N=302$ participants on Prolific, distributed into three conditions: 102 participants in C1, 99 participants in C2 and 101 participants in C3. Participants are selected from the United States, United Kingdom, and Australia and must be fluent in English. Gender-wise, 192 were women, 103 were men, 4 self-specified their gender and 3 declined to state their gender. Age-wise, 94 participants were between Ages 18 and 29, 91 were between Ages 30 and 39, 44 were between Ages 40 and 49, and 73 were over Age 50. All collected data were included in the analysis, as there was no evidence indicating that any participants provided poor-quality data.
    \paragraph{Experiment 2} We recruited $N=95$ participants on Prolific, distributed into three conditions: 30 participants in C1, 34 participants in C2 and 31 participants in C3. Participants are selected from the United States, United Kingdom, and Australia and must be fluent in English. Gender-wise, 52 were women, 41 were men, and 2 declined to state the gender. Age-wise, 38 participants were between Ages 18 and 29, 37 were between Ages 30 and 39, 10 were between Ages 40 and 49, and 10 were over Age 50. Participants in the first study were \textit{not} allowed to participate in the second.

    \subsection{Experiment 1: Quantitative Results}
    \label{sec:experiment1-results}

    \begin{table*}[!ht]
        \centering 
        \setlength{\tabcolsep}{3pt} 
        \sisetup{separate-uncertainty}
        \caption[]{Results of the human experiment. R: Recommendation-driven, O: AI-explanation-only, H: Hypothesis-driven. O-R: Over-Reliance. U-R: Under-Reliance. Lower is better in all measures ($\downarrow$). Winners/significances are indicated in bold.}
        \begin{tabular}{l|c @{\ $\pm$\ } cc @{\ $\pm$\ } cc @{\ $\pm$\ } c|ccc}
        \toprule
        & \multicolumn{2}{c}{R} & \multicolumn{2}{c}{O} & \multicolumn{2}{c}{\highlight{H}} & R vs. O  & R vs. H & O vs. H \\
        \midrule
        Time (min) & $\mathbf{17.92}$ & $\mathbf{9.34}$ & 18.62 & 11.26 & 18.09 & 9.26 & $p=0.808, r=0.02$ & $p=0.993, r=0.00$ & $p=0.892, r=0.01$ \\
    
        Brier score & 0.29 & 0.07 & 0.30 & 0.07 & $\mathbf{0.27}$ & $\mathbf{0.06}$ & $p=0.566, r=0.05$ & $\mathbf{p=0.016, r=0.20}$ & $\mathbf{p=0.003, r=0.25}$ \\
    
        O-R (\%) & 73.86 & 20.91 & 54.21 & 22.51 & $\mathbf{53.30}$ & $\mathbf{22.73}$ & $\mathbf{p<0.001, r=0.45}$ & $\mathbf{p<0.001, r=0.45}$ & $p=0.946, r=0.01$ \\
    
        U-R (\%) & $\mathbf{17.81}$ & $\mathbf{20.35}$ & 41.25 & 27.18 & 24.42 & 18.19 & $\mathbf{p<0.001, r=0.52}$ & $\mathbf{p=0.001, r=0.25}$ & $\mathbf{p<0.001, r=0.39}$ \\
        \bottomrule
        \end{tabular}
        \label{tab:housing-results}
    \end{table*}

    In this section, we show the results of experiment 1, in which we explore whether \textit{hypothesis-driven} can improve task efficiency and task performance, and reduce reliance compared to \textit{recommendation-driven} and \textit{AI-explanation-only}.

    We performed a Shapiro-Wilks test to check the data normality and we found that our data was not normally distributed ($p<0.05$). Therefore, we apply non-parametric Kruskal-Wallis test. We then perform post-hoc Mann-Witney U test to do pairwise comparisons. The results are shown in Table~\ref{tab:housing-results}. The significant differences between the two conditions are indicated in bold where $p < 0.05$.

    \paragraph{Task efficiency} Table~\ref{tab:housing-results} shows the completion time in three conditions. There is no statistically significant difference among these three conditions ($p \approx 0.9$). We reject \textbf{H1a/b}. \textbf{\textit{This shows that hypothesis-driven does not take more time to complete the task than recommendation-driven and AI-explanation-only}}.

    \paragraph{Task performance} We evaluate participants' decision-making performance by using the Brier score. A lower Brier score indicates better decision accuracy. As seen in Table~\ref{tab:housing-results}, \textbf{\textit{participants in the \textit{hypothesis-driven} condition ($M=0.267$, $SD=0.063$) have a lower Brier score than the other two approaches}} (C1: ($M=0.290$, $SD=0.071$), C2: ($M=0.295$, $SD=0.073$)). We \textbf{accept H2a/b} with small effect sizes. Therefore, hypothesis-driven helps participants be confident when they make a correct decision, and be less confident when they make a wrong decision.

    \paragraph{Over-reliance} In Table~\ref{tab:housing-results}, \textbf{\textit{hypothesis-driven ($M=53.30$, $SD=22.73$) reduced over-reliance significantly compared to recommendation-driven ($M=73.86$, $SD=20.91$)}} ($p=1.5 \times 10^{-8}$, $r=0.449$). We \textbf{accept H3a} with a medium effect size of $0.45$. Moreover, \textit{AI-explanation-only} ($M=54.21$, $SD=22.51$) also reduces over-reliance compared to the recommendation-driven approach ($p=1.6 \times 10^{-8}$, $r=0.450$).

    \paragraph{Under-reliance} In Table~\ref{tab:housing-results}, \textit{\textbf{hypothesis-driven ($M=24.42$, $SD=18.19$) significantly reduced under-reliance compared to AI-explanation-only ($M=41.25$, $SD=27.18$)}} ($p=1.09 \times 10^{-6}$, $r=0.387$). We \textbf{accept H3b} with a medium effect size of $0.39$. This is not surprising because we expect that participants in the \textit{AI-explanation-only} condition are the most likely to `under-rely' due to being unable to compare evidence across hypotheses nor see a recommendation. Recommendation-driven ($M=17.81$, $SD=20.35$) has the least under-reliance value because participants were given  recommendations.

    \paragraph{Subjective Questionnaire} We also asked participants to answer a subjective questionnaire after the task. The questionnaire consists of twelve questions to measure \textit{In control}, \textit{Preference}, \textit{Mental demand}, \textit{System complexity} and \textit{Trust}. Details of the questions are provided in \href{https://thaole25.github.io/aij-supp/}{online supplementary materials}.

    \begin{table*}[h]
        \centering 
        \setlength{\tabcolsep}{3pt} 
        \sisetup{separate-uncertainty}
        \caption[]{Results of the subjective questionnaire. R: Recommendation-driven, O: AI-explanation-only, H: Hypothesis-driven. Values are ranged from 0 (disagree strongly) to 10 (agree strongly). Winners/significances are indicated in bold.}
        \begin{tabular}{l|c @{\ $\pm$\ } cc @{\ $\pm$\ } cc @{\ $\pm$\ } c|ccc}
        \toprule
        & \multicolumn{2}{c}{R} & \multicolumn{2}{c}{O} & \multicolumn{2}{c}{\highlight{H}} & R vs. O  & R vs. H & O vs. H \\
        \midrule
        In control ($\uparrow$) & $\mathbf{7.22}$ & $\mathbf{2.44}$ & 5.73 & 2.83 & 7.01 & 2.46 & $\mathbf{p<0.001,r=0.32}$ & $p=0.496,r=0.06$ & $\mathbf{p<0.001,r=0.27}$ \\
        Preference ($\uparrow$) & $\mathbf{5.51}$ & $\mathbf{2.67}$ & 4.05 & 2.79 & 5.41 & 3.03 & $\mathbf{p<0.001,r=0.30}$ & $p=0.930,r=0.01$ & $\mathbf{p=0.002,r=0.26}$ \\
        Mental demand ($\downarrow$) & $\mathbf{3.96}$ & $\mathbf{2.64}$ & 5.41 & 2.68 & 4.54 & 2.80 & $\mathbf{p<0.001,r=0.30}$ & $p=0.168,r=0.11$ & $\mathbf{p=0.028,r=0.18}$ \\
        Complexity ($\downarrow$) & $\mathbf{4.38}$ & $\mathbf{2.53}$ & 5.65 & 2.91 & 4.42 & 2.95 & $\mathbf{p=0.001,r=0.26}$ & $p=0.899,r=0.01$ & $\mathbf{p=0.003,r=0.24}$ \\
        Trust ($\uparrow$) & 4.92 & 1.69 & 3.62 & 1.95 & $\mathbf{5.16}$ & $\mathbf{1.98}$ & $\mathbf{p<0.001,d=0.72}$ & $p=0.352,d=0.13$ & $\mathbf{p<0.001,d=0.79}$ \\
        \bottomrule
        \end{tabular}
        \label{tab:housing-subjective-results}
    \end{table*}

    In Table~\ref{tab:housing-subjective-results}, \textit{AI-explanation-only} is significantly worse than the other conditions in all facets. Moreover, there is no statistical difference between \textit{recommendation-driven} and \textit{hypothesis-driven}. The reason is that we conduct between-subject experiments so each participant has access to only one condition and they do not have another condition to compare to. If we run within-subject experiments to measure subjective questions in the future, participants can compare different decision-making approaches and evaluate which one they prefer the most.

    Notably, the \textit{hypothesis-driven} condition did not result in the highest mental demand. This is because the \textit{AI-explanation-only} condition was more difficult to interpret. In the \textit{AI-explanation-only} condition, participants did not know which hypothesis the evidence was referring to, making it challenging to make sense of the information. Moreover, when participants relied on their prior knowledge, they could check their hypothesis and assess the evidence to confirm it, which is easier than trying to understand a machine recommendation when it differs.

    \subsection{Experiment 2: Qualitative Results}

    \begin{figure}[!ht]
        \centering
        \includegraphics[width=0.8\linewidth]{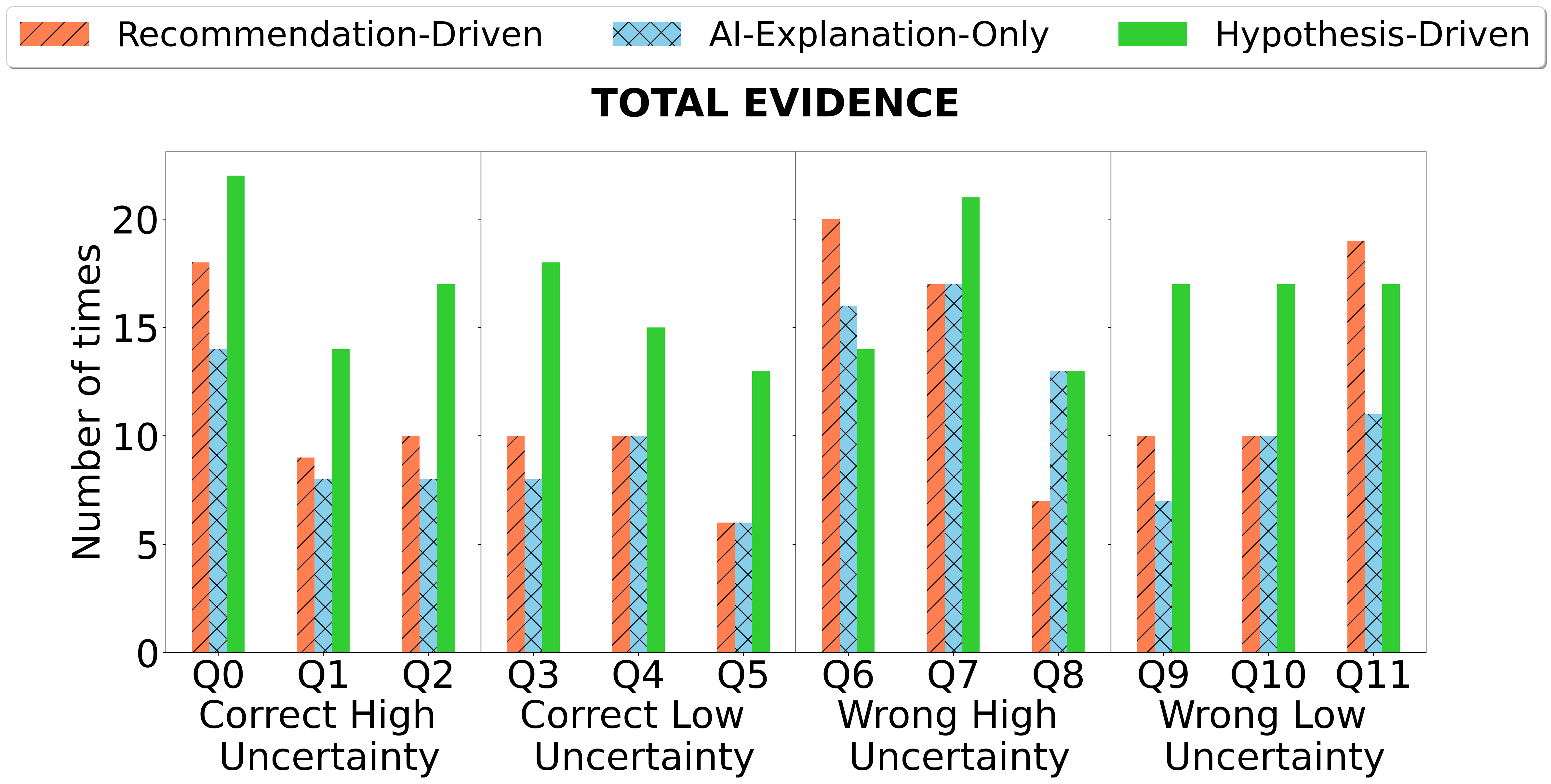}
        \caption{Frequency of using evidence to make a decision.}
        \label{fig:evidence}
    \end{figure}

    \begin{figure}[!ht]
        \centering
        \includegraphics[width=0.8\linewidth]{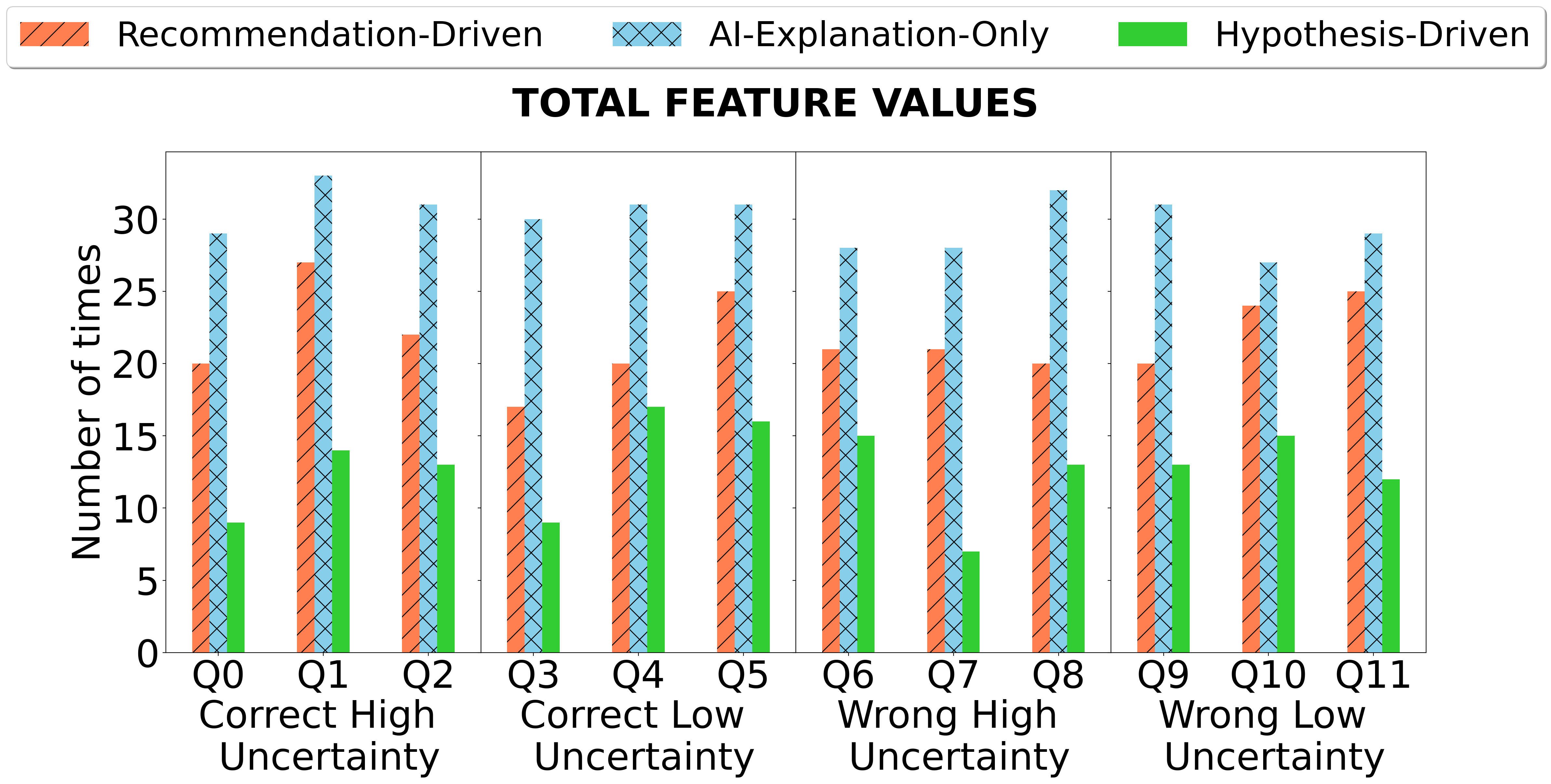}
        \caption{Frequency of using feature values to make a decision.}
        \label{fig:features}
    \end{figure}

    In Figures~\ref{fig:evidence} and~\ref{fig:features}, we illustrate the number of times that participants used feature values and evidence to make their decisions based on the text analysis. The questions in this section (Q0-Q11) refer to the 12 house instances used in the experiment described in Section~\ref{sec:housing:conduct}. Participant comments are attributed to the question number, not the participant ID.

    \textbf{\textit{For the recommendation-driven approach, participants use the feature values to confirm whether the decision aid's prediction and explanation are reliable or not}}. If participants think the feature values do not match the evidence explanation, they will go with the feature values to make the final decision. Some examples that the study participants in the recommendation-driven condition go against the decision aid's prediction: 
    \longquote{Here, I believe the decision aid is mistaken. My rating would be medium because the house is very old which is overlooked by the model. Other features are all decent or above decent but the house age is an important feature.}{Q11}
    \longquote{The location of the property is low so I thought that would bring down the price}{Q7}

    Ignoring evidence is, of course, a good strategy if the decision-maker believes that the evidence is wrong. However, \textbf{\textit{recommendation-driven does not help participants to be aware of the high uncertainty among multiple predictions}}. This limitation is mitigated by the hypothesis-driven approach.

    \textbf{\textit{For the AI-explanation-only approach, participants often rely on the feature values and not on the evidence explanation to make a decision}}. This is not surprising because participants can find it difficult to interpret the evidence without seeing the label that the evidence is referred to.
    We attribute this to the cognitive effort to link evidence to hypotheses, leading to participants ignoring evidence and relying on input feature values to make their decisions. This is a noteworthy limitation of \textit{AI-explanation-only} as it makes people overlook the explanation if the link to the evidence is unclear. In the study by \citet{GajosIncidental22}, the link from feature attributions to the task solution is more straightforward than in our study, which may explain the divergence of results.

    \textbf{\textit{Participants more often use the evidence to make a decision in the hypothesis-driven approach}} than in recommendation-driven or AI-explanation-only. This shows participants took advantage of the model evidence. In the two baseline conditions, participants tended to ignore evidence seemingly due to the inability to interpret it, which means they will fail to take advantage of the underlying model. In Figure~\ref{fig:evidence}, there are only two exceptions at Q6 and Q11 where the evidence is not the most used in the hypothesis-driven condition.

    \begin{figure}[!ht]
        \centering
        \includegraphics[width=0.8\linewidth]{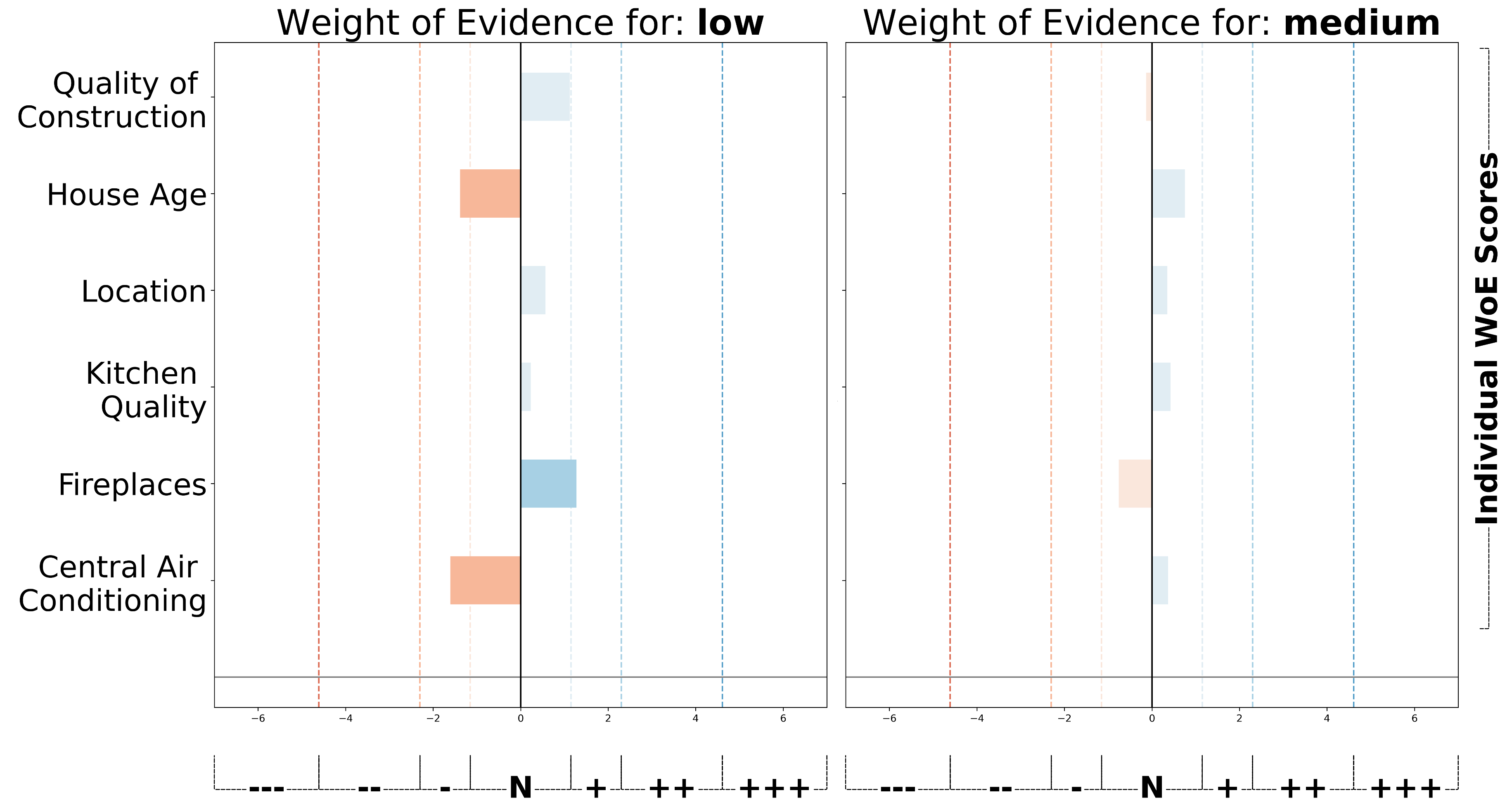}
        \caption{An example of \textbf{\textit{uncertainty awareness}}  (Q6).}
        \label{fig:c3-uncertainty-awareness}
    \end{figure}

    \begin{figure}[!ht]
        \centering
        \includegraphics[width=0.8\linewidth]{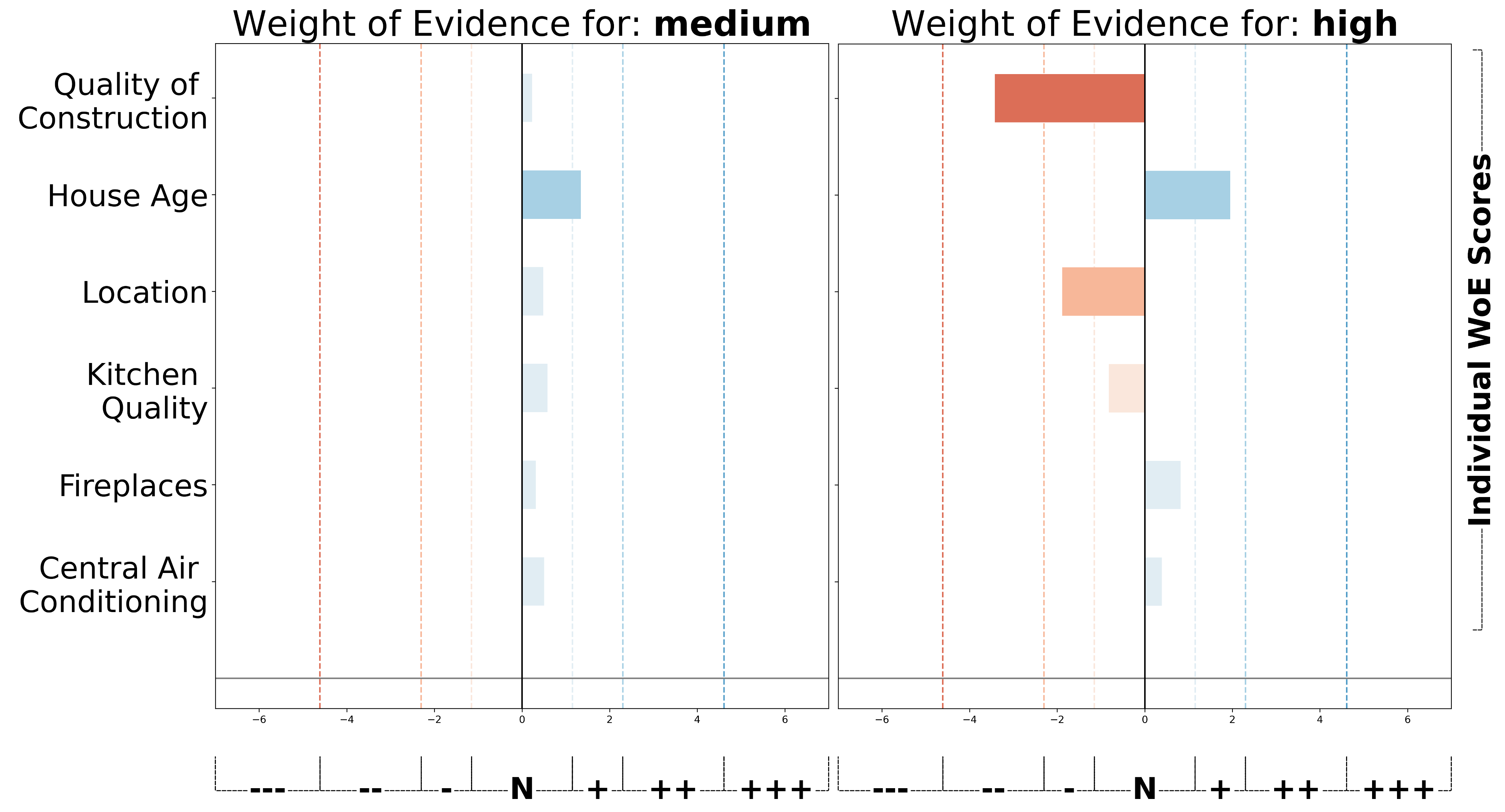}
        \caption{An example of \textbf{\textit{deceptive evidence}} (Q9).}
        \label{fig:c3-deception-unawareness}
    \end{figure}

    We found that in hypothesis-driven, participants reported that it was difficult to make decisions for two main reasons: 
    \begin{itemize}
        \item \textbf{\textit{Uncertainty awareness}}: This is where there are multiple hypotheses with similar strength evidence. Participants are aware of the uncertainty in the model solely based on the positive and negative evidence provided for all hypotheses. In this case, participants use the input feature values or choose the hypothesis that they think is slightly better than the others when making the final decision. Figure~\ref{fig:c3-uncertainty-awareness} shows an example where two hypotheses \textit{low} and \textit{medium} both have a positive and negative weight of evidence, especially in the top three important features. For instance, some participants explicitly explain their uncertainty in the text as follows.
        \longquote{I was choosing between high and medium. Quality of construction, age and location are the most important features. When it was high, these were all positive. Kitchen quality and fireplaces were negative, but these are not as important.}{Q0}
        \longquote{The amount of negative or positive evidence for low or medium is the same, including the three more important factors. Both medium or low could be viable but medium has less variance and is overall more balanced.}{Q6}
        \longquote{The house is clearly not in a high bracket, but it is somewhat difficult to decide between low and medium. There are stronger indicators in low, going both ways, while medium has largely insignificant indicators. Low has a significant negative weighting for house age and this pushed me towards medium.}{Q6}
        
        \item \textbf{\textit{Deceptive evidence}}: When the evidence was strongest for an incorrect option. In this case, many participants just follow the evidence and make the wrong decision. Figure~\ref{fig:c3-deception-unawareness} illustrates an example of Q9 where we have all positive evidence in hypothesis \textit{medium}, but strong negative evidence in hypothesis \textit{high}. Therefore, all participants choose hypothesis \textit{medium}, but hypothesis \textit{high} is the ground truth. Future work will need to address the challenge of building trustworthy evidence.
    \end{itemize}

    In summary, the qualitative analysis showed that participants took advantage of the decision aid more in the hypothesis-driven condition than in recommendation-driven and explanation-only conditions. Further, we also found that participants recognised model uncertainty in the hypothesis-driven condition. However, there still remains a limitation of having deceptive evidence.

%% file: skin-domain.tex
\section{Study 2: Visual Evaluative AI for Skin Cancer Diagnosis}
\label{sec:skin-domain}
In this section, we will describe the implementation of \textit{Visual Evaluative AI} in a case study of supporting skin cancer diagnosis. The goal of this human experiment is to understand the differences in terms of decision accuracy, decision time and user satisfaction between the recommendation-driven and hypothesis-driven approaches in supporting skin cancer diagnosis. We tested only these two conditions because the study follows a within-subject design that requires participants to complete all conditions. Adding another condition would make the study too long for participants and could lead to fatigue. Moreover, these two conditions are the main approaches in decision support systems that we aim to compare.

    \subsection{Basic Concepts in Skin Cancer Diagnosis}
    Dermatologists usually diagnose skin cancer by following ABCD rule~\cite{Nachbar94} or 7-point checklist criteria~\cite{Argenziano98,kawahara2018seven}. Comparing these two criteria, the 7-point checklist gives higher sensitivity, which is the accuracy of correctly identifying malignant lesions~\cite{Argenziano04}. In this section, we provide an overview of the basic concepts in skin cancer diagnosis by focusing on the 7-point checklist criteria.

    Following the terminology in Table 2 (page 18)~\cite{Kittler16} and \cite{Braun05}, the seven concepts used in the 7-point checklist are: (1) atypical pigment network, (2) blue-white veil, (3) atypical vascular pattern, (4) irregular streaks, (5) irregular pigmentation, (6) irregular dots/globules, and (7) regression structures. This is a scoring system that assigns a score to each criterion, and the total score is used to classify the lesion as benign or malignant. Details of the scoring are described in Table~\ref{tab:7pt_criteria}. Based on the 7-point checklist above, \citet{kawahara2018seven} provide a 7-point criteria evaluation database called \textit{Derm7pt} dataset. Using this dataset, we can extract 12 concepts included in Table~\ref{tab:12_concepts}. Each concept can indicate whether the lesion is benign or malignant.

    \begin{table}[width=.55\linewidth,cols=3,pos=ht]
        \caption{7-point checklist criteria}
        \label{tab:7pt_criteria}
        \begin{tabular*}{\tblwidth}{@{} c|cc@{} }
        & \textbf{Criteria} & \textbf{7-point score}\\
        \toprule
        \multirow{3}{*}{Major criteria} & Atypical pigment network & 2 \\
        & Blue-white veil & 2 \\
        & Atypical vascular pattern & 2 \\
        \midrule
        \multirow{4}{*}{Minor criteria} & Irregular streaks & 1 \\
        & Irregular pigmentation & 1\\
        & Irregular dots/globules & 1\\
        & Regression structures & 1\\
        \bottomrule
        \end{tabular*}
    \end{table}

    \begin{table}[width=.6\linewidth,cols=2,pos=ht]
        \caption{12 concepts used in the supervised method~\cite{Yan23towards}}
        \label{tab:12_concepts}
        \begin{tabular*}{\tblwidth}{@{} c|c@{} }
        Concept Name  & Description \\
        \midrule
        \textit{Atypical Pigment Network} & concept activation for melanoma \\
        \textit{Typical Pigment Network} & concept activation for benign \\
        \midrule
        \textit{Blue Whitish Veil} & concept activation for melanoma \\
        \midrule
        \textit{Irregular Vascular Structures} & concept activation for melanoma \\
        \textit{Regular Vascular Structures} & concept activation for benign \\
        \midrule
        \textit{Irregular Pigmentation} & concept activation for melanoma \\
        \textit{Regular Pigmentation} & concept activation for benign \\
        \midrule
        \textit{Irregular Streaks} & concept activation for melanoma \\
        \textit{Regular Streaks} & concept activation for benign \\
        \midrule
        \textit{Regression Structures} & concept activation for melanoma \\
        \midrule
        \textit{Regular Dots and Globules} & concept activation for benign \\
        \textit{Irregular Dots and Globules} & concept activation for melanoma \\
        \bottomrule
        \end{tabular*}
    \end{table}
    
    Despite there being two common diagnostic outputs (benign and malignant), there are seven classes of skin lesions that can be diagnosed based on the HAM10000 dataset~\cite{Tschandl18}, as shown in Table~\ref{tab:7_classes}. Each class has different characteristics and is classified into either benign or malignant. It is worth noting that there is no clear answer for \textit{actinic keratoses}~\footnote{Based on a conversation with Prof. Peter Soyer}. This class can be considered as a pre-cancerous lesion, which can be classified as either benign or malignant. In Australia, actinic keratoses are common and usually treated as benign. However, in other countries, they can be considered as malignant. In this paper, we follow the authors of the HAM10000 dataset, which classified actinic keratoses as malignant~\cite{Tschandl20}.

        \begin{table}[width=.4\linewidth,cols=2,pos=ht]
            \caption{Seven output classes}\label{tab:7_classes}
            \begin{tabular*}{\tblwidth}{@{} c|l@{} }
                Diagnosis & Lesion Type \\
                \midrule
                \multirow{5}{*}{Benign}
                & (BKL) Benign keratosis-like lesions \\
                & (DF) Dermatofibroma \\
                & (NV) Melanocytic nevi  \\
                & (VASC) Vascular lesions \\
                \midrule
                \multirow{2}{*}{Malignant}
                & (AKIEC) Actinic keratoses \\
                & (BCC) Basal cell carcinoma \\
                & (MEL) dermatofibroma \\
            \end{tabular*}
        \end{table}

    \subsection{Dataset and Model Implementation}
    We use the \textbf{HAM10000 dataset}~\cite{Tschandl18} to train all models (original CNN backbones, ICE, ICE+WoE, PCBM and PCBM+WoE). This dataset has a total of 10015 dermoscopic images and seven output classes: Actinic keratoses (AKIEC), basal cell carcinoma (BCC), benign keratosis (BKL), dermatofibroma (DF), melanoma (MEL), melanocytic nevi (NV) and vascular lesion (VASC). Among these seven classes, actinic keratoses (AKIEC), basal cell carcinoma (BCC), and melanoma (MEL) are malignant, while benign keratosis (BKL), dermatofibroma (DF), melanocytic nevi (NV) and vascular lesion (VASC) are benign. We choose the HAM10000 dataset instead of the 7-point checklist dataset~\cite{kawahara2018seven} (2000 images) because HAM10000 is a larger dataset with more samples, which can help achieve more accurate classifiers. HAM10000 is also more generalised, as it was collected from multiple institutes, whereas the 7-point checklist dataset was collected from a single source. Lastly, HAM10000 is more well-known, with many more existing works and baseline models using this dataset.
    
    We balanced the dataset by applying Weighted Random Sampler~\footnote{\url{https://pytorch.org/docs/stable/_modules/torch/utils/data/sampler.html}} and data augmentation. Finally, each class has 1000 samples that were used for the training process, making a total of 7000 samples for seven classes. The test set is selected as a fraction of the original dataset (without augmentation). As in the original HAM10000, class DF has the lowest number of samples (i.e., 75 samples). Therefore, we choose 20 samples in each class for the test set, which represents 26\% of class DF. We then have a total of 140 samples (20 samples $\times$ 7 classes) for the test set to evaluate the model performance.

    \begin{table}[width=.6\linewidth,cols=3,pos=h]
        \caption{The number of positive and negative samples for each concept in the concept bank.}
        \label{tab:concept-bank}
        \begin{tabular*}{\tblwidth}{@{} lcc@{} }
        \toprule
        Concept Name & Positive Samples & Negative Samples \\
        \midrule
        Atypical Pigment Network      & 230              & 781              \\
        Typical Pigment Network       & 381              & 630              \\
        Blue Whitish Veil             & 195              & 816              \\
        Irregular Vascular Structures & 71               & 940              \\
        Regular Vascular Structures   & 117              & 894              \\
        Irregular Pigmentation        & 305              & 706              \\
        Regular Pigmentation          & 118              & 893              \\
        Irregular Streaks             & 251              & 760              \\
        Regular Streaks               & 107              & 904              \\
        Regression Structures         & 253              & 758              \\
        Irregular Dots and Globules   & 448              & 563              \\
        Regular Dots and Globules     & 334              & 677 \\
        \bottomrule             
        \end{tabular*}
    \end{table}

    Since images in the HAM10000 dataset do not have the concept labels, to get the concept labels for the PCBM model, we train Concept Activation Vectors (CAVs)~\cite{Kim18} on the \textbf{7-point checklist dataset}~\cite{kawahara2018seven} to obtain the concept library $I$. Followed the previous work~\cite{yuksekgonul2023posthoc,Yan23towards}, we have 12 concepts: \textit{Atypical Pigment Network}, \textit{Typical Pigment Network}, \textit{Blue Whitish Veil}, \textit{Irregular Vascular Structures}, \textit{Regular Vascular Structures}, \textit{Irregular Pigmentation}, \textit{Regular Pigmentation}, \textit{Irregular Streaks}, \textit{Regular Streaks}, \textit{Regression Structures}, \textit{Irregular Dots and Globules} and \textit{Regular Dots and Globules}. The PCBM model then used the trained CAVs based on these 12 concepts and applied that to extract the concept. In more detail, we show the number of positive and negative samples for each concept based on the 7-point checklist dataset in Table~\ref{tab:concept-bank}, which is used to learn the concept embeddings in the PCBM model. The number of positive samples refers to how many images contain the concept, while the number of negative samples refers to how many images do not contain the concept. In our final chosen model, for each concept, we select 50 positive samples (contain the concept) and 50 negative samples (do not contain the concept). The learning rate was set to $0.01$ and ridge regression was used at the classifier layer of PCBM.

    \subsection{Computational Experiments}
    \label{sec:evaskan:computational}
    We evaluate the computational performance of the combined models ICE+WoE and PCBM+WoE on the skin cancer dataset (HAM10000)~\cite{Tschandl18}. We compare them in terms of accuracy and investigate the impact of the number of concepts on the performance of ICE+WoE.

    We apply different CNN backbones (Resnet50, ResneXt50 and  Resnext152) to train the original CNN models, ICE, ICE+WoE, PCBM and PCBM+WoE. Based on each CNN backbone, we compare the original backbone model with ICE, ICE+WoE, PCBM and PCBM+WoE. We use the F1-score metric to evaluate the performance of the models. The F1-score is calculated as the harmonic mean of precision and recall, which is defined as:

    \begin{equation*}
        \text{Precision} = \frac{\text{True Positive (TP)}}{\text{True Positive (TP)} + \text{False Positive (FP)}}
    \end{equation*}

    \begin{equation*}
        \text{Recall} = \frac{\text{True Positive (TP)}}{\text{True Positive (TP)} + \text{False Negative (FN)}}
    \end{equation*}

    \begin{equation*}
        F1 = 2 \times \frac{precision \times recall}{precision + recall}
    \end{equation*}

    where True Positive (TP) is the number of correctly predicted malignant cases; False Positive (FP) is the number of incorrectly predicted malignant cases; True Negative (TN) is the number of correctly predicted benign cases; False Negative (FN) is the number of incorrectly predicted benign cases.

    \begin{figure}[!ht]
        \noindent\begin{minipage}[b]{0.49\linewidth}
            \centering
            \includegraphics[width=\linewidth]{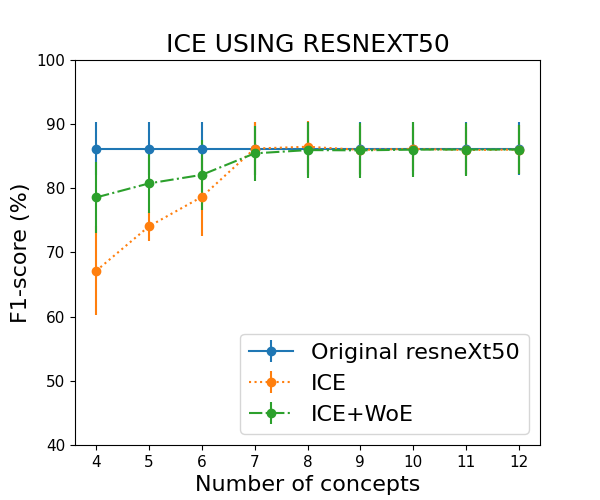}
        \end{minipage}
        \noindent\begin{minipage}[b]{0.49\linewidth}
            \centering
            \includegraphics[width=\linewidth]{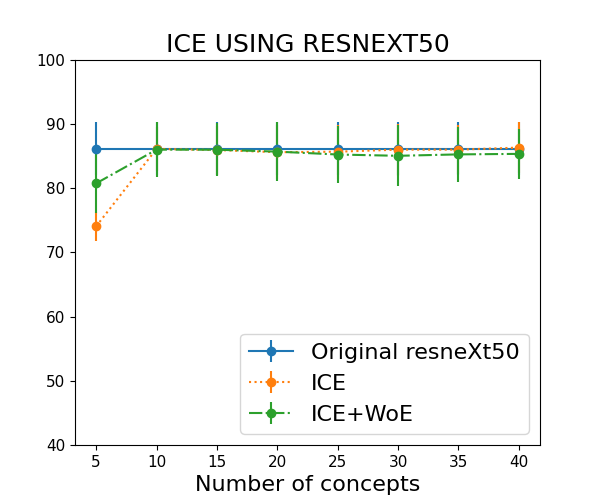}
        \end{minipage}
        \caption{F1-score of ICE, ICE+WoE and the original ResneXt50 over different number of concepts. The left figure shows the performance of ICE and ICE+WoE with a small number of concepts (4-12), while the right figure shows the performance of ICE and ICE+WoE with a larger range of number of concepts (5-100).}
        \label{fig:rx50-by-concept}
    \end{figure}

    \begin{figure}[!ht]
        \centering
        \includegraphics[width=0.49\linewidth]{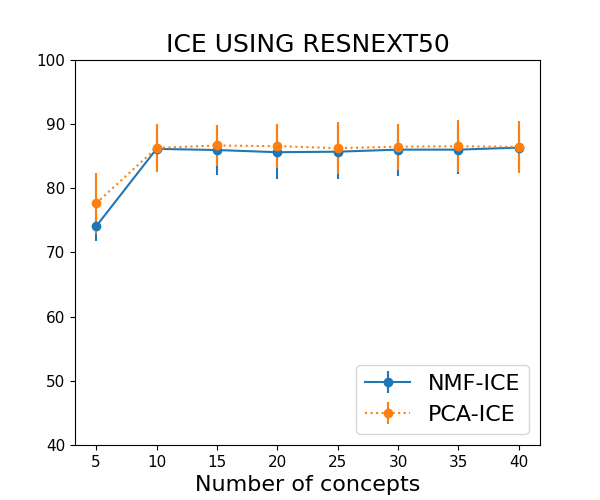}
        \caption{Comparing different reducers NMF and PCA.}
        \label{fig:rx50-by-reducer}
    \end{figure}

    \begin{table}[!ht]
        \centering
        \begin{tabular}{lllll}
        \toprule
        CNN & Model & Precision $\uparrow$ & Recall $\uparrow$ & F1-Score $\uparrow$ \\
        Backbone & & & &\\
        \midrule
        \multirow{5}{*}{Resnet50}
        & Backbone & $83.08 \pm 5.98$ & $85.33 \pm 6.20$ & $\mathbf{84.04 \pm 5.01}$ \\
        & ICE(7) & $73.34 \pm 8.69$ & $87.50 \pm 10.04$ & $78.99 \pm 4.91$ \\
        & \highlight{ICE(7)+WoE} & \highlight{$80.13 \pm 5.44$} & \highlight{$82.00 \pm 6.81$} & \highlight{$80.85 \pm 4.55$} \\
        & PCBM(12) & $73.93 \pm 8.94$ & $82.08 \pm 12.67$ & $76.58 \pm 6.31$ \\
        & \highlight{PCBM(12)+WoE} & \highlight{$80.73 \pm 5.21$} & \highlight{$84.25 \pm 3.35$} & \highlight{$82.32 \pm 2.98$} \\
        \midrule
        \multirow{5}{*}{ResneXt50}
        & Backbone & $85.46 \pm 4.63$ & $87.25 \pm 6.31$ & $86.20 \pm 4.18$ \\
        & ICE(7) & $84.23 \pm 5.49$ & $88.58 \pm 5.41$ & $\mathbf{86.20 \pm 4.11}$ \\
        & \highlight{ICE(7)+WoE} & \highlight{$84.73 \pm 5.00$} & \highlight{$86.33 \pm 4.76$} & \highlight{$85.45 \pm 4.25$} \\
        & PCBM(12) & $78.93 \pm 8.28$ & $83.17 \pm 14.43$ & $79.83 \pm 8.28$ \\
        & \highlight{PCBM(12)+WoE} & \highlight{$84.48 \pm 4.86$} & \highlight{$85.50 \pm 3.98$} & \highlight{$84.92 \pm 3.64$} \\
        \midrule
        \multirow{5}{*}{Resnet152}
        & Backbone & $84.49 \pm 6.48$ & $86.08 \pm 5.70$ & $\mathbf{84.96 \pm 3.09}$ \\
        & ICE(7) & $78.30 \pm 8.11$ & $87.42 \pm 7.48$ & $82.10 \pm 4.37$ \\
        & \highlight{ICE(7)+WoE} & \highlight{$81.21 \pm 4.90$} & \highlight{$85.08 \pm 5.14$} & \highlight{$83.01 \pm 4.13$} \\
        & PCBM(12) & $76.49 \pm 7.75$ & $87.08 \pm 5.15$ & $81.09 \pm 4.21$ \\
        & \highlight{PCBM(12)+WoE} & \highlight{$82.97 \pm 5.37$} & \highlight{$84.83 \pm 4.04$} & \highlight{$83.73 \pm 2.99$} \\
        \bottomrule
        \end{tabular}
        \caption{Performance for the original CNN model, ICE, ICE+WoE, PCBM and PCBM+WoE. The ICE model uses an NMF (non-negative matrix factorization) reducer. ICE(7) represents the ICE model with 7 different concepts. PCBM(12) is the PCBM model with 12 labelled concepts. \textit{mean} $\pm$ \textit{standard deviation} of the performance are reported over 20 random seeds. Winners are indicated in bold.}
        \label{tab:evaskan-computational}
    \end{table}

    \paragraph{\textbf{ICE+WoE and PCBM+WoE achieved comparable performance to the original CNN models}} 
    Table~\ref{tab:evaskan-computational} reports the performance of ICE(7), ICE(7)+WoE, PCBM(12) and PCBM(12)+WoE using three different CNN backbone models (Resnet50, Resnet152~\cite{Kaiming16} and ResneXt50~\cite{Saining17}). We select 12 concepts for PCBM based on previous work~\cite{yuksekgonul2023posthoc,Yan23towards}. For ICE, we run experiments with the number of concepts ranging from 5 to 40. As shown in Figure~\ref{fig:rx50-by-concept}, performance peaks at 7 concepts. Therefore, the final comparison in this table is made between ICE(7) and PCBM(12).
    
    The results show that ICE(7)+WoE and PCBM(12)+WoE achieve comparable performance to the original CNN models. Particularly, with ResneXt50, the F1-score of ICE(7)+WoE and PCBM(12)+WoE are $85.45 \pm 4.25$ and $84.92 \pm 3.64$, respectively, while the original ResneXt50 has an F1-score of $86.20 \pm 4.18$. Therefore, ICE(7)+WoE (using 7 features) and PCBM(12)+WoE (using 12 features) show comparable performance compared to the original ResneXt50 with 2048 features. Moreover, similar to the findings in~\cite{zhang2021improving}, when we compare the performance using different reducers as in Figure~\ref{fig:rx50-by-reducer}, NMF and PCA (principal component analysis), PCA provided the best performance but could be less interpretable compared to NMF.
 
    \paragraph{\textbf{Having more concepts did not lead to better accuracy}} 
    Figure~\ref{fig:rx50-by-concept} shows the performance of the original ResneXt50, ICE and ICE+WoE over different numbers of concepts from 5 concepts to 40 concepts. Two figures from the left show the performance of ICE using the NMF reducer. When there are 5 concepts, ICE(5)+WoE ($80.43 \pm 4.60$) has a significantly higher F1-score than ICE(5) ($74.08 \pm 2.26$) ($p=2.41 \times 10^-6 < 0.001$, $d=1.753$). Since we have 2048 features at the classifier layer of ResneXt50, ResneXt50 outperforms ICE(5)+WoE and ICE(5) significantly ($p<0.001$). But the performance of both ICE+WoE and ICE match the performance of the original ResneXt50 when we have at least 7 concepts. Particularly, with \textit{as few as 7 concepts}, ICE and ICE+WoE achieve similar performance to the original ResneXt50 using 2048 features. The performance of ICE and ICE+WoE also stopped improving at 7 concepts with a backbone of ResneXt50. The reason is that when we apply a reducer in ICE (e.g. NMF), some important concepts are detected at first. Then after we increase the number of concepts, some noisy concepts are detected, which could lead to a slight drop in the performance. Eventually, all important concepts are found and match the performance of the original CNN model.

    In summary, the results show that with a few number of concepts (i.e., 7 concepts), we can achieve comparable performance compared to the original CNN models. Therefore, this indicates the accuracy of the evidence being generated, which is potentially useful to the decision-makers. Importantly, despite the concept-based models (ICE(7), ICE(7)+WoE, PCBM(12) and PCBM(12)+WoE) being slightly less accurate than the CNN backbones, it would also be much easier for users to interpret and evaluate the evidence by not showing too many concepts.

    \subsection{Example Concept-Based Explanations}

    \begin{figure}[!ht]
        \centering
        \includegraphics[width=\linewidth]{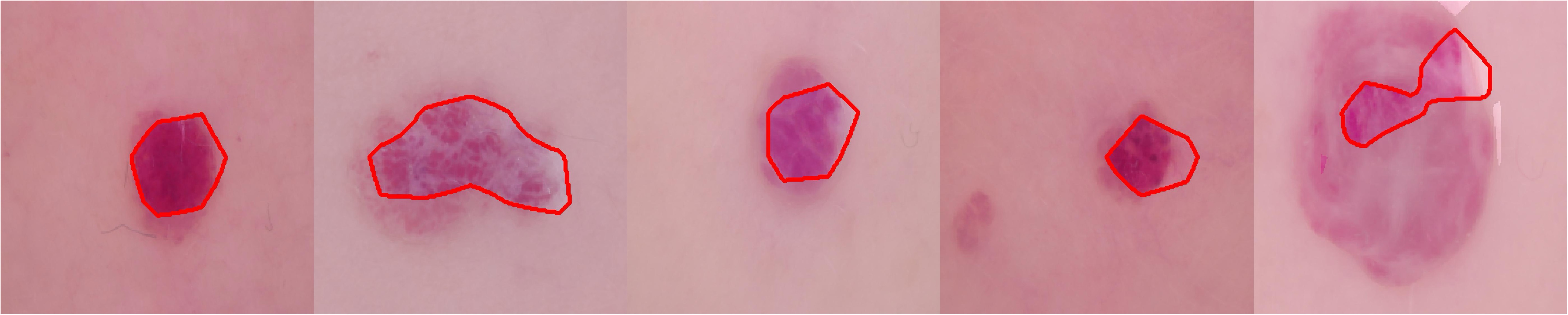}
        \caption{Reddish structures}
        \label{fig:feat1-vascular}
    \end{figure}

    \begin{figure}[!ht]
        \centering
        \includegraphics[width=\linewidth]{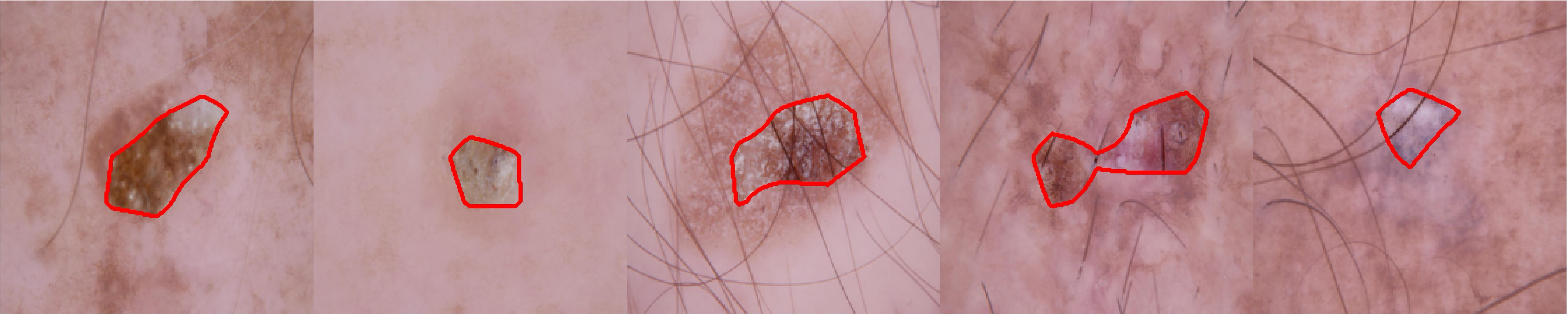}
        \caption{Lines (Hair)}
        \label{fig:feat7-lines}
    \end{figure}

    In preparation for the human study, we iteratively try different numbers of concepts and work closely with a domain expert to refine and label the concepts. The domain expert is an academic dermatologist with more than 40 years of experience in the field. We decided to use the ICE+WoE model to generate concept-based explanations for our human study because currently the labelled concepts found by the PCBM+WoE model are often wrong and considered unreliable by the expert. One possible reason could be the Out-of-Distribution (OOD) problem, as the concept bank is trained on the 7-point checklist dataset~\cite{kawahara2018seven} and then applied to the HAM10000 dataset~\cite{Tschandl18}. Furthermore, ICE+WoE also achieves higher performance than PCBM+WoE in our computational experiments in Section~\ref{sec:evaskan:computational}. Therefore, it is important to note that the concepts in our human experiment are not labelled by the model, but are labelled by the domain expert.
    
    \cref{fig:feat1-vascular} is an example of the concept \textit{Reddish structures}. There are a total of seven concepts used in the human study, including (1) Reddish structures, (2) Irregular pigmentation, (3) Irregular dots and globules, (4) Whitish veils, (5) Irregular pigmentation, (6) Dark irregular pigmentation and (7) Lines (Hair), which can be found in \href{https://thaole25.github.io/aij-supp/}{online supplementary materials}. Each concept is represented by five examples (instances) in the training set that segment areas of interest, specifically, the red polygon outlines. These outlines are segmented using the ICE method, not hand-drawn. These examples are selected by getting the best feature importance, which can be estimated using the method in TCAV~\cite{Kim18}. In this case, we select five best examples that have the highest feature importance. We also ensure that the lesions are different from each other among the examples. The examples are arranged in ascending order of feature importance scores, from left to right. For instance, although we could not find a concept name that matches all five examples in Figure~\ref{fig:feat7-lines}, we chose to annotate it as \textit{lines (hair)} because the three rightmost examples, i.e., the ones with the highest scores, focus on the hair in the images.
    
    These seven concepts are identified as important by the classifier to make a decision. We then use them to generate the evidence by applying the WoE~\cite{Melis21} and conduct a human experiment. Study participants can decide whether to use the evidence generated by the model to make the final decision. For example, the model detects \textit{lines} (Figure~\ref{fig:feat7-lines}) as an important concept, though it is a \textit{confounding feature} and should be ignored in making the diagnosis.

    \subsection{Study Design}

    \begin{figure}[!ht]
        \centering
        \includegraphics[width=0.8\linewidth]{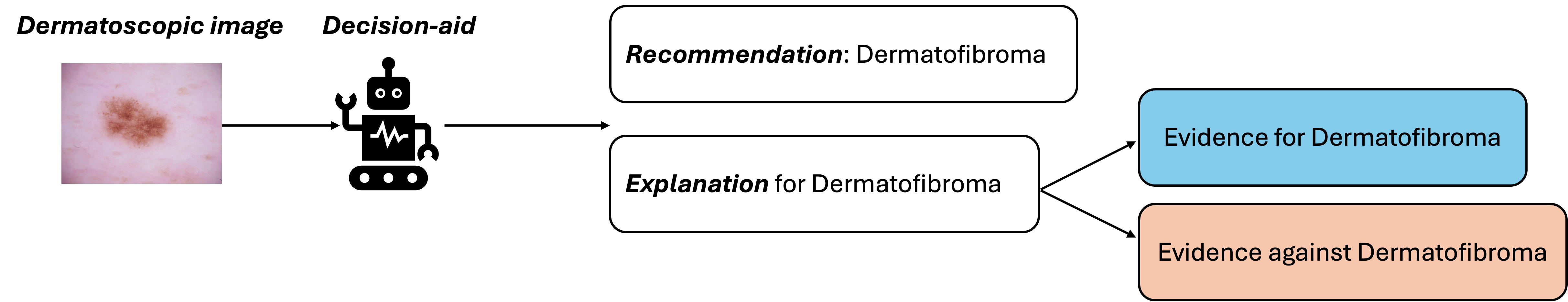}
        \caption{The recommendation-driven flow}
        \label{fig:recommendation-flow}
    \end{figure}

    \begin{figure}[!ht]
        \centering
        \includegraphics[width=0.8\linewidth]{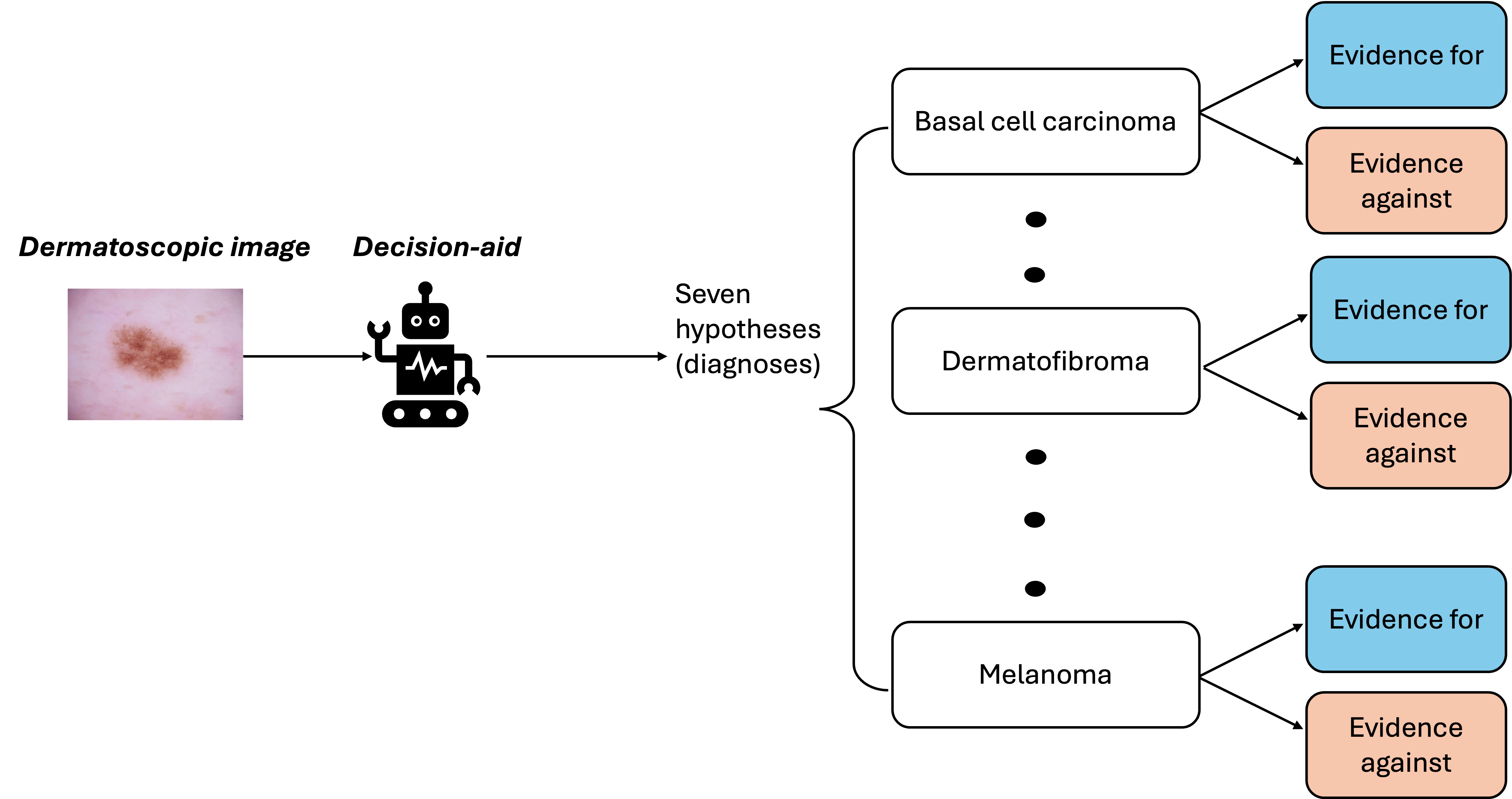}
        \caption{The hypothesis-driven flow}
        \label{fig:hypothesis-flow}
    \end{figure}

    \begin{figure}[!ht]
        \centering
        \includegraphics[width=\linewidth]{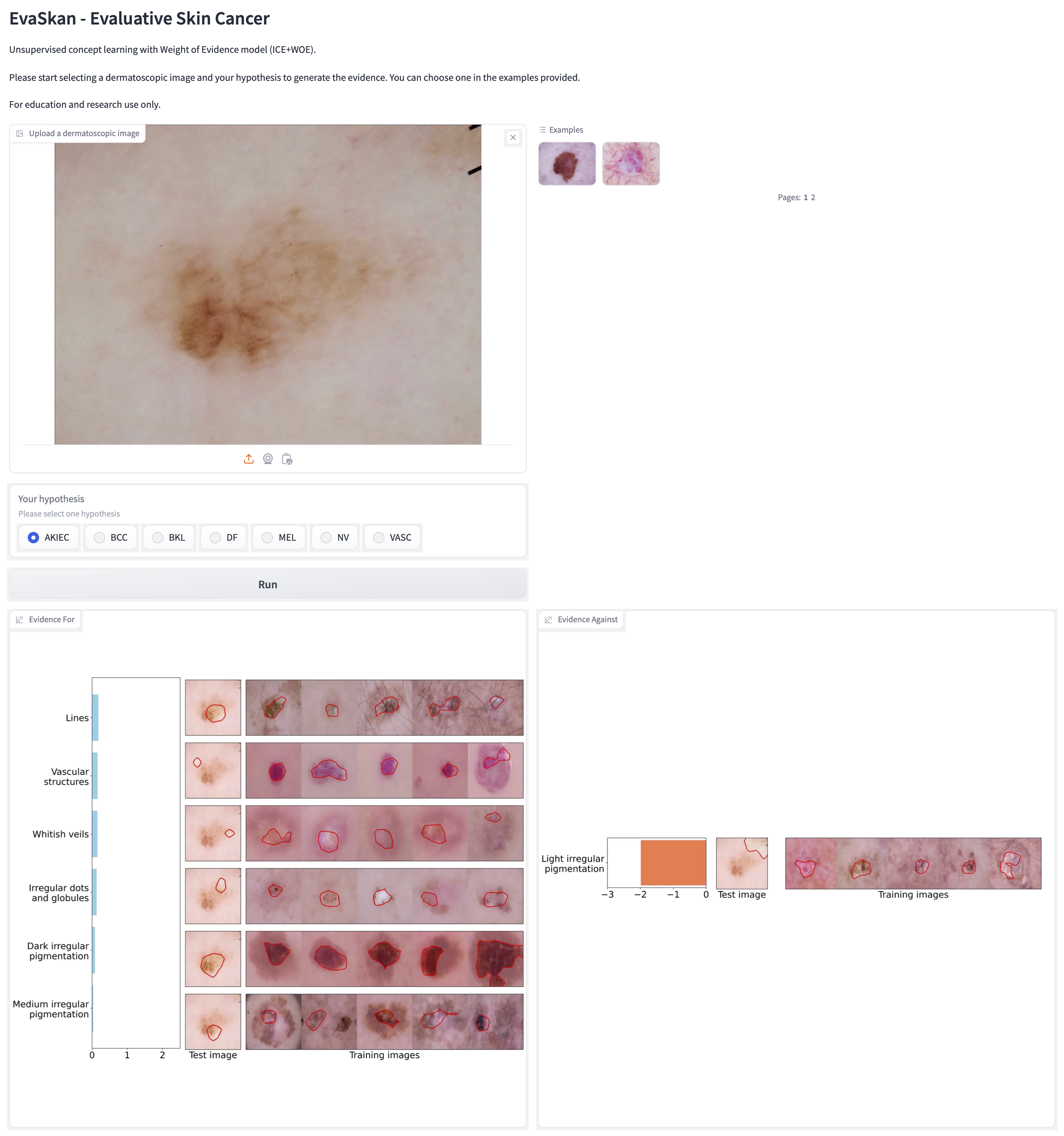}
        \caption{A screenshot of the EvaSkan web application}
        \label{fig:evaskan-demo}
    \end{figure}
    
    The study is conducted on a web application called \textit{EvaSkan} (Evaluative Skin Cancer), where participants are asked to make skin cancer diagnoses using the two approaches (recommendation-driven and hypothesis-driven). An example EvaSkan interface is shown in Figure~\ref{fig:evaskan-demo}. A short paper about an earlier version of EvaSkan is available here~\cite{le2024visual}. To see the full protocol of our experiment and the web interfaces, please refer to~\href{https://thaole25.github.io/aij-supp/}{online supplementary materials}.

    We formed an interdisciplinary team of AI researchers and a skin cancer expert. The AI researcher team prepared the drafted materials for the user study. Then, the domain expert, who is a co-author and an academic dermatologist with more than 40 years of experience in the field, provided feedback to improve these materials during iterative discussions. The domain expert also did a pilot study before conducting the main study to ensure the materials were ready for participants.

    There are three phases in the study. In phase 1, we collect the demographic information of the participants including their roles, background, years of experience, and whether they are familiar with AI and skin cancer diagnosis. In phase 2, we conduct a within-subject design study. Study participants use two web interfaces for two conditions (\textit{recommendation-driven} and \textit{hypothesis-driven}) and perform skin cancer diagnosis tasks on a web page. In the \textit{recommendation-driven} condition, participants are given the AI prediction for the skin cancer diagnosis and the explanation of that prediction. In the \textit{hypothesis-driven} condition, participants are given explanations for and against all possible hypotheses. The web page records the interaction log of participants.

    There are a total of sixteen different tasks (questions) for two conditions, eight of which are in one condition, and eight in the other. Sixteen questions are uniformly distributed into four categories: (1) where the model gives correct predictions with high uncertainty, (2) where the model gives correct predictions with low uncertainty, (3) where the model gives wrong predictions with high uncertainty and (4) where the model gives wrong predictions with low uncertainty.
    
    Conditions and tasks are randomly counterbalanced. Specifically, the order of the conditions is randomised and the order of the tasks within a condition is randomised. Further, out of sixteen images, we also randomly select images for each condition to minimise selection bias of using the same set of images in each condition. At the end of phase 2, we ask them to evaluate their preferences in these two conditions using bipolar scale questions as follows: 
    \begin{enumerate}
        \item In control: Scale these conditions based on how much you are in control of the decision-making process.
        \item Decision-making: Scale these conditions based on how helpful it is to you to make the diagnosis.
        \item Ease of use: Scale these conditions based on how easy it is to use.
        \item Error detection: Scale these conditions based on how easy it is to spot mistakes in the decision aid.
    \end{enumerate}

    Finally, in Phase 3, we conducted a semi-structured interview by asking participants to reflect on how they made the diagnoses in Phase 2 using a think-aloud protocol and open questions about the design of our decision aids (DAs). The study was pre-registered~\cite{leosf2024} and received ethics approval (ID: 23208) before data collection. This study requires a maximum of one hour to finish.

    \subsection{Study Hypotheses}
    Our research hypotheses are as follows:
    \begin{itemize}
        \item \textbf{H1}: The hypothesis-driven approach will take more time to make decisions than the recommendation-driven approach.

        \item \textbf{H2}: The hypothesis-driven approach will help study participants make more accurate decisions than the recommendation-driven approach.
        
        \item \textbf{H3}: Study participants will be more satisfied with the hypothesis-driven approach than the recommendation-driven approach. More specifically,
        \begin{itemize}
            \item \textbf{H3a}: Participants feel they have more control of the decision-making process when using the hypothesis-driven approach compared to the recommendation-driven approach.
            \item \textbf{H3b}: Participants feel the hypothesis-driven approach is more helpful in making a diagnosis than the recommendation-driven approach.
            \item \textbf{H3c}: Participants find the hypothesis-driven approach is easier to use than the recommendation-driven approach.
            \item \textbf{H3d}: Participants find it is easier to spot mistakes in the decision-aid when using the hypothesis-driven approach compared to the recommendation-driven approach.
        \end{itemize}
    \end{itemize}

    \subsection{Participants}

    \begin{table}[width=.76\linewidth,cols=5,pos=ht]
        \caption{Study participant's details. \textit{Year of Experience} refers to the years they have spent in that role.}
        \label{tab:participants}
        \begin{tabular*}{\tblwidth}{llrll}
        \toprule
        ID & Role & Years of & Research/Work & Experience in  \\
        & & Experience & Field & Skin Cancer \\
        & & & & Diagnosis \\
        \midrule
        P0 & PhD Student \& & 5 & Dermatology & No \\
        & Senior Research Technician & & & \\
        P1 & PhD Student & 2 & Melanoma & No\\
         &  &  &  Detection & \\
        P2 & PhD Student \& & 5 & Melanoma & No \\
        & Research Assistant &  & Detection & \\
        P3 & Resident Doctor & 2 & Cutaneous & Yes \\
        &  &  & Phenotyping & \\
        P4 & Melanographer & 1 & Melanography & No\\
        P5 & Resident Doctor & 2 & Dermatology & Yes \\
        P6 & Melanographer & 10 & Dermatology & Yes\\
        P7 & Postdoc & 12 & AI Implementation & No\\
         &  &  & in Skin Cancer & \\
        P8 & Postdoc & 2 & Biostatistics & No\\
        P9 & Principal House Officer & 1 & Dermatology & Yes\\
        P10 & Resident Doctor & 3 & Melanoma Prognosis & Yes\\
        P11 & Senior House Officer & 1 & Dermatology & Yes\\
        P12 & Principal House Officer & 2 & Dermatology & Yes \\
        P13 & Senior House Officer & 3.5 & Melanoma & Yes\\
        \bottomrule
        \\
        \end{tabular*}
    \end{table}

    We recruit individuals with a background in skin cancer through our professional networks. Participants receive 25 AUD upon completing the study. There are a total of \textbf{14} participants whose details are summarised in Table~\ref{tab:participants}.  Gender-wise, there are \textbf{7} females and \textbf{7} males.

    Some participants have used AI decision support tools in research settings, such as \textit{Canfield Dermoscopy Explained Intelligence (DEXI)}, \textit{Canfield's Dermagraphix}, \textit{FotoFinder MoleAnalyzer} and \textit{Lesion Change Detection model}. None of them have used AI tools in clinical settings. We classify participants into either \textit{experienced} or \textit{inexperienced} as in Table~\ref{tab:participants} (\textit{Experience in Skin Cancer Diagnosis}). Experienced participants are individuals who have received clinical training (e.g., resident doctors, senior house officer, principal house officer, senior melanographer~\footnote{\url{https://www.careers.health.qld.gov.au/medical-careers/career-structure}}). Inexperienced participants include PhD students and postdoctoral researchers working in the skin cancer field, but are not specifically trained to become clinicians.

    \subsection{Experiment Variables}
    This study has two independent variables (two \textit{conditions}): recommendation-driven and hypothesis-driven. To compare between these two conditions, we took the following measures:
    \begin{enumerate}
        \item \textit{Time spent on each task (Instance time)}: We measure the time participants spend on each diagnosis task in Phase 2 (one out of sixteen tasks in total). Then, we also calculate the total time spent on all tasks in one condition (\textit{Total time});
        \item \textit{Brier score} We measure the effectiveness of task performance by evaluating both the confidence of the participant and the correctness of the answer. The formula is:
        \begin{equation}
            \text{BS}_{p} = \frac{1}{N} \sum_{i=1}^{N}(C_{p,i} - A_{p,i})^2 
        \end{equation}
        where: $C_{p,i}$ is the confidence level of participant $p$ in question $i$, ranging from 0 to 1; $A_{p,i}$ is the answer score of participant $p$ in question $i$, either 0 (wrong answer) or 1 (right answer). $N$ is the number of questions, which is $N=8$ in one condition;
        \item \textit{Selected hypotheses}: In the hypothesis-driven condition, we record which hypotheses are being checked. We then calculate the \textit{percentage of selected hypotheses} by dividing the number of selected hypotheses by the total number of hypotheses, which is 7. This measure can indicate whether study participants used their prior knowledge in the decision-making process;
        \item \textit{Self-reported bipolar scales}: We ask participants to evaluate their preferences in these two conditions in terms of \textit{in control}, \textit{decision-making}, \textit{ease of use} and \textit{error detection} using bipolar scale questions as in the study design.
    \end{enumerate}

    \subsection{Quantitative Results}

    We now show the performance and self-reported bipolar scales of participants when interacting with two conditions, recommendation-driven and hypothesis-driven.

    \subsubsection*{Performance}

    \begin{figure}[!ht]
        \centering
        \includegraphics[width=0.8\linewidth]{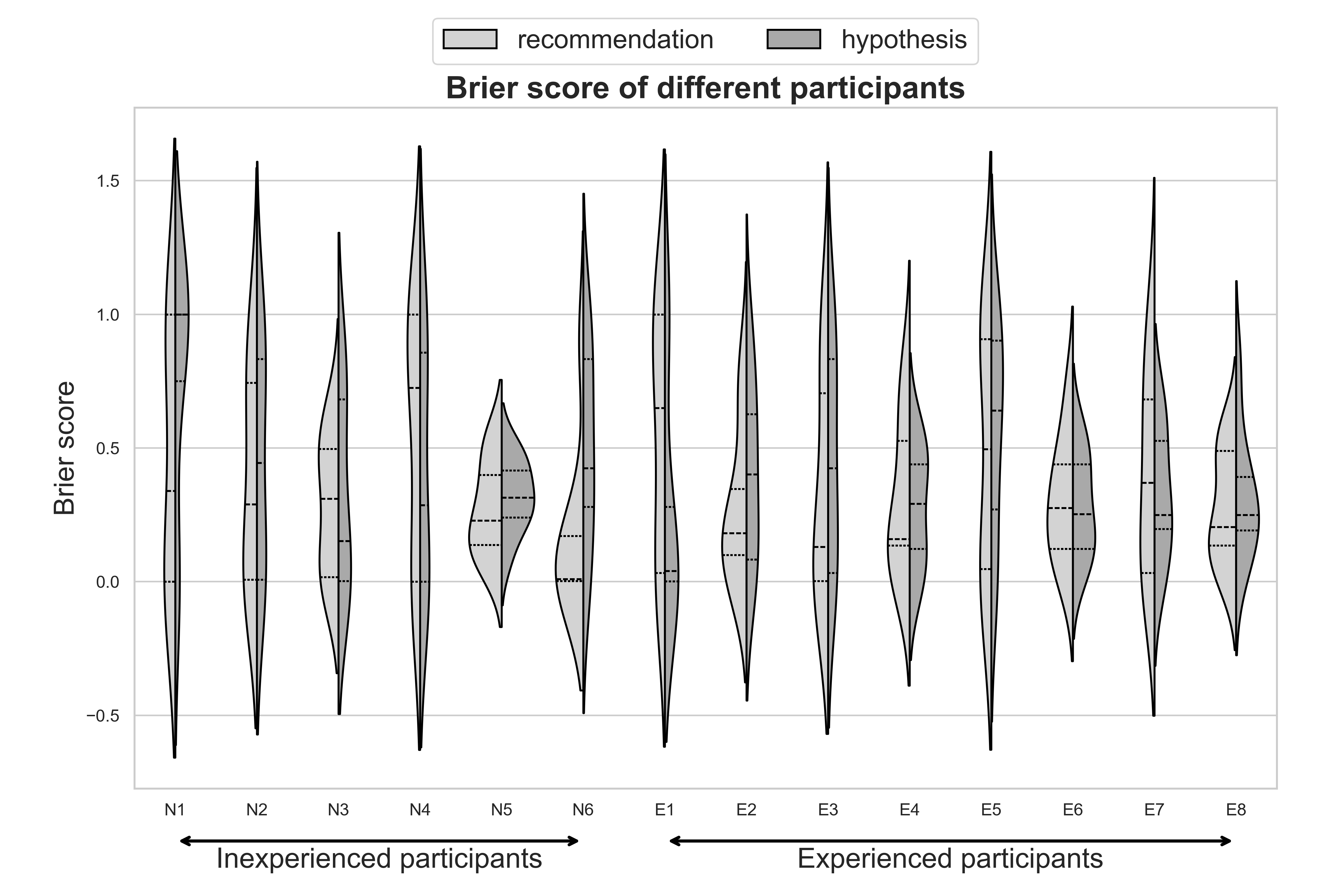}
        \caption{Brier score for all participants. The IDs here are different from the IDs in Table~\ref{tab:participants} to protect participants' privacy. Participants are separated into \textit{Experienced participants} and \textit{Inexperienced participants} based on the classification in \textit{Experience in Skin Cancer Diagnosis} in Table~\ref{tab:participants}.}
        \label{fig:violin-brier-score}
    \end{figure}

    \begin{table}[width=.9\linewidth,cols=5,pos=ht]
        \caption[]{Performance of participants in terms of time required to complete all tasks and the Brier score. \textit{R}: Recommendation-driven, \textit{H}: Hypothesis-driven. Winners/significances are indicated in bold.}
        \label{tab:results-quantitative}
        \begin{tabular*}{\tblwidth}{p{0.15\linewidth}p{0.15\linewidth}|r @{\ $\pm$\ } rr @{\ $\pm$\ } rr}
        \toprule
        & & \multicolumn{2}{c}{R} & \multicolumn{2}{c}{H} & R vs. H (Wilcoxon t-test) \\
        \midrule
        \multirow{3}{*}{All} & Total time (s) $\downarrow$ & \textbf{449.78} & \textbf{301.95} & 580.68 & 310.91 & $\mathbf{p=0.05, r=0.53}$\\
        & Instance time (s) $\downarrow$ & \textbf{56.22} & \textbf{45.59} & 72.59 & 48.71 & $\mathbf{p < 0.001, r=0.37}$\\
        & Brier score $\downarrow$ & \textbf{0.36} & \textbf{0.36} & 0.40 & 0.36 & $p=0.35, r=0.09$\\
        \midrule
        \multirow{2}{*}{Experienced} & Total time (s) $\downarrow$ & \textbf{503.21} & \textbf{377.06} & 618.11 & 369.43 & $p=0.31, r=0.36$ \\
        & Instance time (s) $\downarrow$ & \textbf{62.90} & \textbf{53.96} & 77.26 & 55.05 & $\mathbf{p=0.005, r=0.35}$\\
        & Brier score $\downarrow$ & 0.37 & 0.35 & \textbf{0.36} & \textbf{0.32} & $p=0.94, r=0.01$ \\
        \midrule
        \multirow{2}{*}{Inexperienced} & Total time (s) $\downarrow$ & \textbf{378.54} & \textbf{165.38} & 530.78 & 234.57 & $p=0.09, r=0.68$ \\
        & Instance time (s) $\downarrow$ & \textbf{47.32} & \textbf{29.38} & 66.35 & 38.37 & $\mathbf{p=0.006,r=0.40}$\\
        & Brier score $\downarrow$ & \textbf{0.35} & \textbf{0.38} & 0.46 & 0.40 & $p=0.16, r=0.20$ \\
        \bottomrule
        \end{tabular*}
    \end{table}

    Participants' performances are summarised in Table~\ref{tab:results-quantitative}. Regarding the time spent to complete the task, the hypothesis-driven condition takes more time than the recommendation-driven condition for all participants significantly. Moreover, experienced participants take more time to evaluate each instance (task) in both interfaces, suggesting that they are more careful in making decisions. Moreover, for experienced participants, the distribution of Brier scores is tighter in the hypothesis-driven, with most participants achieving scores close to 0. This result implies that experienced participants performed better with the hypothesis-driven, whereas inexperienced participants had better performance with the recommendation-driven interface. Note that there is no significant difference in the Brier score between the two conditions for all participants. However, the time required to complete all tasks is very close to the significance level ($p=0.05$) between the two conditions for all participants. Based on the result, we can accept \textbf{H1} (time) and reject \textbf{H2} (accuracy).

    \subsubsection*{Subjective Bipolar Scales}

    \begin{figure}[!ht]
        \centering
        \includegraphics[width=0.8\linewidth]{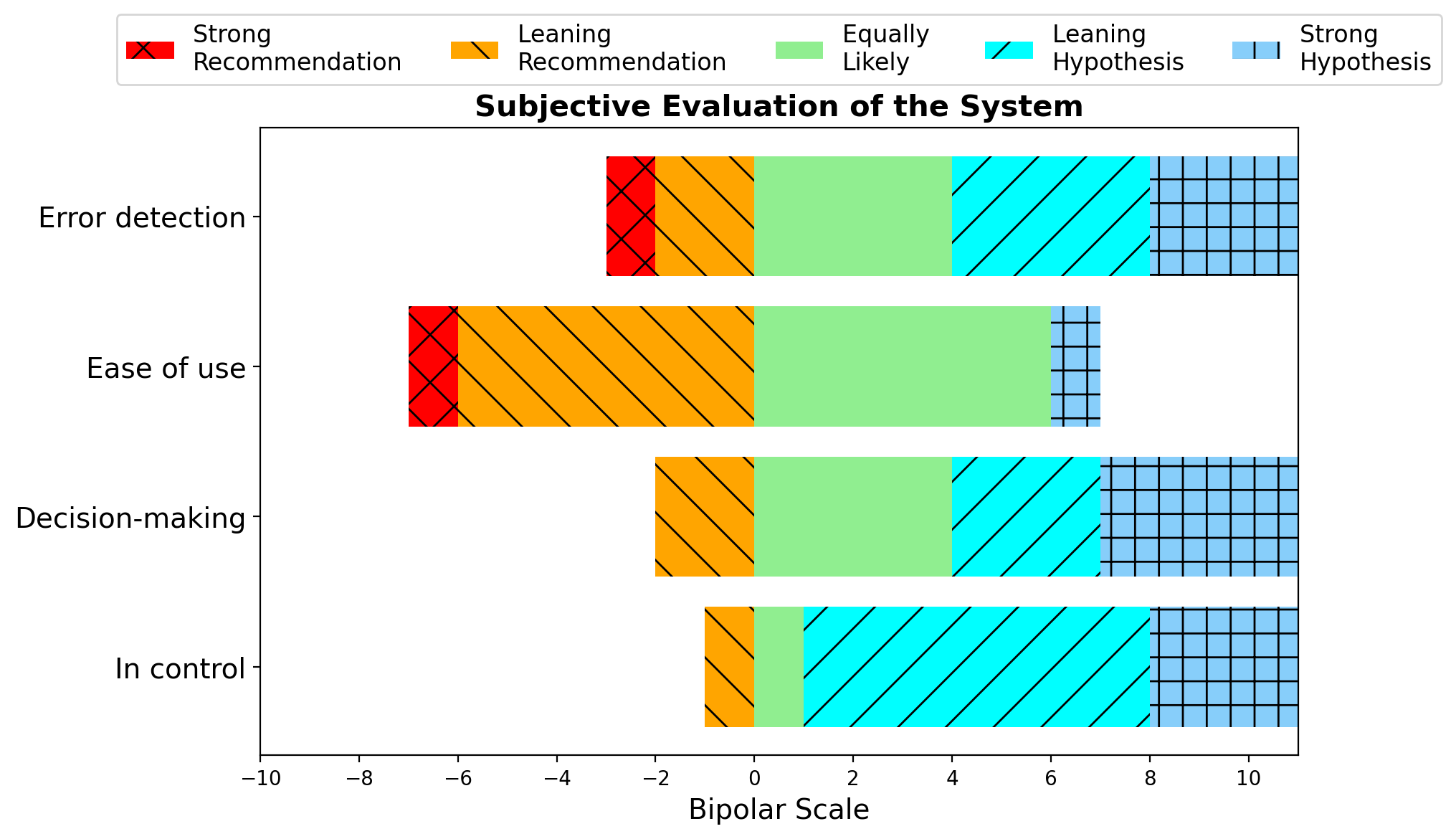}
        \caption{Bipolar scale counts of the approach's subject metrics.}
        \label{fig:bipolar-scale}
    \end{figure}

    \begin{table}[width=.5\linewidth,cols=3,pos=ht]
        \caption[]{Results of Subjective Bipolar Scales (-5 = Recommendation-driven is the best; 0 = equally likely, 5 = Hypothesis-driven is the best). Significances are indicated in bold.}
        \label{tab:results-bipolar}
        \begin{tabular*}{\tblwidth}{l|c @{\ $\pm$\ } c|c}
        \toprule
        Metric & \multicolumn{2}{|c|}{Mean $\pm$ std} & One sample t-test \\
        \midrule
        In control & 2.71 & 1.82 & $\mathbf{p<0.001, d=1.49}$\\
        Decision-making & 1.57 & 2.53 & $\mathbf{p=0.037, d=0.62}$\\
        Ease of use & -1.07 & 2.27 & $p=0.100, d=0.472$\\
        Error detection & 0.93 & 2.79 & $p=0.234, d=0.333$\\
        \bottomrule
        \end{tabular*}
    \end{table}

    From Figure~\ref{fig:bipolar-scale} and Table~\ref{tab:results-bipolar}, participants show preferences in the hypothesis-driven interface in terms of \textit{in control}, \textit{decision-making} and \textit{error detection}. However, the two interfaces have no significant difference in \textit{ease of use}, with a slight preference for the recommendation-driven interface. This result is consistent with the quantitative results in Table~\ref{tab:results-quantitative} where the recommendation-driven interface is faster to complete. We can accept \textbf{H3a} (in control) and \textbf{H3b} (helpful in decision-making), but reject \textbf{H3c} (ease of use) and \textbf{H3d} (error detection).

    \subsubsection*{Selected Hypotheses}
    \begin{figure}[!ht]
        \centering
        \includegraphics[width=0.8\linewidth]{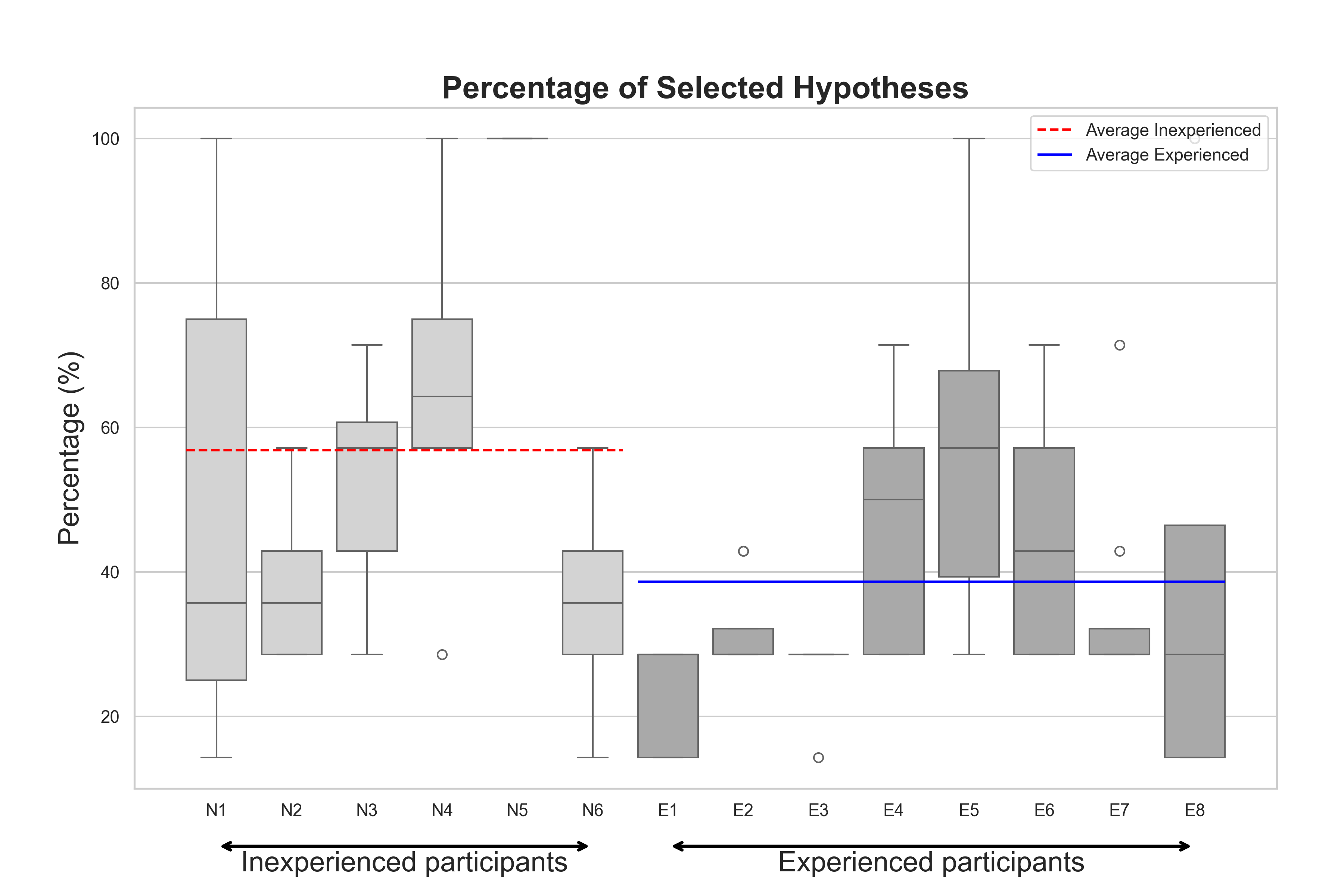}
        \caption{Percentage of selected hypotheses by participants.}
        \label{fig:percentage-hypotheses}
    \end{figure}

    Figure~\ref{fig:percentage-hypotheses} shows the percentage of selected hypotheses by participants in the hypothesis-driven condition. Experienced participants tend to select fewer hypotheses than inexperienced participants when checking the evidence. This result implies that experienced participants apply their thoughts about the diagnosis and only check the evidence for the hypotheses they think are relevant. Inexperienced participants, on the other hand, tend to check more hypotheses as they do not have the base knowledge to decide the possible hypotheses by themselves.

    \subsection{Qualitative Results}

    In this section, we discuss the interview results after participants experience the two interfaces. The results show that they are consistent with previous literature. More importantly, the qualitative results provide insights into the differences between the recommendation-driven and the hypothesis-driven interfaces, which have not been addressed before.

    \subsubsection{Perceived Accuracy and Reliability of the Decision Support}

    Most participants commented that it was difficult to say about the accuracy because they did not know about the ground truth of the test lesions. A participant said that their decision would be very different without the AI information, so it would be helpful in the future to evaluate whether there is a difference between having a decision support and without having a decision support.

    \longquote{I am pretty sure having this information would be a great help for me as a beginner.}{inexperienced user}

    \longquote{A suggested diagnosis can distract me from making my own decision [...] I tried to look at the image first and then look at the diagnosis provided by the AI.}{inexperienced user}

    Regarding whether the AI can help participants make better or worse decisions, a participant said \textbf{AI could make them doubt their initial decision} if the AI contradicts their decision. However, \textbf{it would be helpful if they were not sure and wanted a second opinion from the AI}. Should the AI disagree with the participant's opinions, they would go with their own diagnosis.

    \longquote{It really depends on the case. If you see a clinical image and if you are very sure that it's [a diagnosis], I think the AI is going to make you doubt your initial thought if the AI contradicts your decision. But I think it would be helpful in case you are not sure and you would want a second opinion from an AI system.}{inexperienced user}

    \subsubsection{User Concerns}
    A participant expressed concern regarding the \textbf{dataset} used to train the AI, noting that the training data performs extremely well within a controlled environment. The real-world data can be variable and they would be keen to know how AI would behave outside of the training data. Moreover, HAM10000 data do not have a category between nevus and melanoma. They look similar but very separate. This is the current failure of the training dataset that needs to be addressed by differentiating this middle ground in the future.
    
    Regarding the \textbf{described features (concepts)}, they are high-level, which can be difficult for inexperienced participants. A participant suggested it would be more helpful if there were more descriptions of the labels (e.g., what \textit{reddish structures} and \textit{pigmentations} mean). Users also suggested that the concepts (which are represented as segmentations) should focus on the lesion, and have the ability to ignore the irrelevant background (e.g., rulers, dark corners, etc.), which are confounding features. But in fact, this shows that explainable AI is helpful in terms of helping participants take into consideration if they should trust the AI or not.

    \longquote{Sometimes the AI picked up something from the background rather than the lesion itself, so I think it would give a false recommendation.}{inexperienced user}

    \longquote{When you see that the model is picking up things outside of the lesion of interest, then you know that the diagnosis is going to be skewed by that. [...] But I think that helps you as a clinician [...]. And you're thinking about whether should I change my diagnosis to be more in line with the model or should I stick with my own.}{experienced user}

    Regarding the \textbf{test images}, some of them are straightforward but some others are more challenging. This shows the limitation of the current study as doctors would need more information such as patient history, clinical location, age, sex, other images, etc. before they can make a diagnosis rather than relying on a single dermoscopic image.

    \longquote{The pictures you selected [...] Some of them were more quite straightforward but some of them were more challenging, particularly the pigmented lesions were often at the borderline. The recommendation-driven struggles with those.}{experienced user}

    \subsubsection{Perceived Evidence Quality}
    Overall, the provided evidence is quite good. Participants used both the concepts on the images and the weight of evidence of each concept. They often began by examining the segmentations that highlight specific features to assess their reliability -- essentially evaluating how much they agreed with them. Based on this assessment, they either incorporated the weight of evidence into their diagnostic process or relied more heavily on their own dermoscopic knowledge. If they think the evidence was reliable, they would calibrate their decision to align more closely with the decision aid.
    \longquote{[..] realise that the areas that it was highlighting weren't necessarily the area of interest. So I sort of go more with my own understanding of dermoscopy and think what would I say if I didn't have the aid.}{inexperienced user}
    \longquote{If it pulled out a good region of interest and it has what I would assume is the correct amount of weight put into it. Then that would sway more than if you're looking at something that you don't think is relevant.}{experienced user}

    Sometimes, the segmentation on the test lesion and the training lesions are not consistent in finding the same area on the skin. For instance, in one example, the segmentation on the test lesion shows the whole lesion, including some external skin. But in the example training lesions, it just points at a specific area within the larger lesion. In another case, the evidence overlaps in multiple features - highlighting the same area but representing different features.

    \subsubsection{Use of Additional Self-Sourced Evidence in Decision-Making}
    Some participants tried to look at the original lesion first before checking the recommendation and evidence provided by the AI. Participants with experience in the skin cancer field would apply their knowledge to validate whether they should trust the decision aid's provided information. Furthermore, they also found other evidence that the model failed to detect in the explanation. However, an experienced participant said that they did not use external evidence often in the hypothesis model. The self-sourced evidence only influenced probably 15\% of their decision-making, and they mainly relied on the evidence being provided, especially weights of evidence. Overall, the two interfaces have the advantage of pointing out important aspects of the dermoscopy that doctors might have missed.

    \subsubsection{Pros and Cons of the Recommendation-driven Interface}

    The recommendation-driven has the main advantage of \textbf{requiring a shorter time} to do the diagnosis task and simpler to follow the information on the interface. Furthermore, the recommendation-based approach provides the most likely diagnosis, which is assumedly closer to the ground truth by inexperienced participants. Therefore, beginners in the field prefer the recommendation-driven because it is \textbf{more streamlined} and they only need to confirm if they want to follow the recommendation or not.

    \longquote{If you look at lots of lesions, like hundreds of lesions. I think the benefit of the recommended is it sort of directs you [...] It's directing you to the most likely and showing you the evidence. So be a lot quicker.}{experienced user}

    However, a major disadvantage of the recommendation-driven is that it \textbf{can bias the clinicians}. Experienced participants suggested that users should be required to make their own decisions before looking at the AI recommendation. Furthermore, when the recommendation contradicts the user's own diagnosis, despite being very confident in their initial diagnosis, users can doubt their opinions and eventually a wrong AI recommendation can lead to a wrong direction.

    \longquote{Having a decision-aid's recommendation would be a disadvantage because it may influence your own clinical assessment in a negative way.}{inexperienced user}

    \longquote{I'm a huge fan of the recommendation one because as soon as I open the slide, it's just got a diagnosis there. So it's sort of pre-primed you to think what it is.}{experienced user}

    Some experienced participants actually found that the recommendation-based approach easier to make the decision compared to the hypothesis-driven approach because they only have to compare between the AI advice and their own diagnosis, which requires processing less information and is \textbf{clearer than the hypothesis-driven modality}.

    \longquote{I assume it's pulling from a probability-weighted diagnosis [...], so it gives you the most amount of contrast between yourself and the computer, which I like.}{experienced user}

    \longquote{I think it [the recommendation-driven] was clearer than the hypothesis-driven modality in that I found the compare and contrast a bit more finicky to use, whereas the recommendation put forward what the AI thought was the best but you could still use your own decision making when it came to.}{experienced user}

    \subsubsection{Pros and Cons of the Hypothesis-driven Interface}

    Some participants felt more confident in using the hypothesis-driven because it has \textbf{more options} for the users to choose from and they can compare and contrast between different hypotheses. It lets the users check where the AI is looking for each diagnosis. This design is especially important when users disagree with the model's recommendation and want to see the alternative options with their corresponding evidence. However, it \textbf{requires users to have a good base of knowledge} for all seven diagnoses to make comparisons between different hypotheses.

    \longquote{I think the hypothesis-driven interface works better because you can compare between different diagnoses and how the AI looked for in both diagnoses.}{inexperienced user}

    \longquote{I think the hypothesis-driven was a lot more helpful because a lot of that stems from where the recommended viewing parts of the lesion are.}{experienced user}

    When using this interface, participants spent a substantial amount of time evaluating the lesion before checking the evidence. This can \textbf{reduce the bias} of relying on the AI model.

    \longquote{The advantage is that you're still relying on your own initial clinical knowledge first because you're selecting what you think the possible lesions are.}{experienced user}

    However, \textbf{having too much information} can be a disadvantage, especially for participants who have a few years of experience in this field. When there is a lot of information, participants can \textbf{feel uncertain} about which diagnosis they should choose, and they eventually go with their own clinical assessment to make the final decision.

    \longquote{If you have too much information displayed at once, then it becomes hard to pick and choose. But maybe that's for the best if our initial diagnosis is showing as not great evidence, [...] maybe I should reevaluate my own choice.}{experienced user}

    \longquote{It's depending on the experience of the user, whether they are resident or consultant, it's very easy to just click through all of the links and experience decision fatigue as a result}{experienced user}

    Regarding which model is easier to use, it really depends on the setting. If we had individuals who do not have much experience in diagnosis, they would opt for the recommendation model. But for participants who have more experience, would prefer the hypothesis-driven. An experienced participant in the field commented that if they used the aid in a clinical setting, they would be more likely to use the hypothesis model because it allows them to put forward their hypothesis first and avoid bias.

    \subsubsection{Suggested Improvements for the Decision Support}

    Participants have suggested some improvements for the decision aid to help them make better decisions. The first suggestion is about \textbf{improving feature descriptions}. An inexperienced user said adding more detailed descriptions for the features' labels would be useful for them. For example, adding explanations to describe what \textit{reddish structures} mean, etc. and users can hover over the feature's label to see more details.

    Secondly, users suggested adding \textbf{more supporting information} to the aid. Hypothesis-driven could be improved by providing the rank of the hypotheses according to the AI model, such as clearly ordering the hypotheses based on the most likely to the least likely. Alternatively, providing a probability distribution of all possible diagnoses can be very helpful. However, we should do that after the hypotheses have been selected to avoid biasing the initial decision of the user. Furthermore, when giving a recommendation, we should also supply the probability that the model has given to that diagnosis. This can help users to be aware of the certainty of the AI model. Overall, adding a certainty level for each diagnosis for both interfaces is an important improvement.

    \longquote{I didn't like that I had to click through all of them to see all of the evidence for each one. [...] someone maybe with less experience would have to click through all of them and that might be more time-consuming}{inexperienced user}

    Another suggestion is that the decision aid needs to assess the \textbf{chance of malignancy}. This could involve displaying features that suggest whether a lesion is benign or malignant, and explicitly indicating the presence or absence of such characteristics. Additionally, incorporating contextual information such as more images of the surrounding area and the \textbf{lesion's history} can be very helpful. Another idea is to provide \textbf{case-based explanations} for each diagnosis. For example, if the model identifies a lesion as melanoma, it should present visually similar cases with confirmed melanoma diagnoses. Finally, regarding the decision-making workflow, the recommendation-driven could be improved further by implementing the \textbf{human-first workflow}.
    
    \longquote{I like the idea that you look at it and you make up your own mind and it tells you the recommendation afterwards. Whereas if it tells you what it thinks it is before you even look at it, I think there's cognitive guidance happening there.}{experienced user}

%% file: 5-discussion.tex
\section{Discussions}
\label{sec:discussion}
    In this section, we summarise the findings from the experiments and discuss the implications of the results. We also discuss the validity, limitations of the studies and suggest future work.
    \subsection{The Two Sides of the Coin: Recommendation-driven and Hypothesis-driven}

    \subsubsection{Housing price prediction domain}
    Participants using the hypothesis-driven approach required a similar time to complete the task compared to the recommendation-driven approach. Participants in the hypothesis-driven condition also made higher quality decisions than \textit{recommendation-driven} and \textit{AI-explanation-only} based on the Brier score. The results indicated that the \textit{hypothesis-driven} gave study participants a more complete picture of the underlying decision aid than the other two approaches, helping them to make use of the AI models when they are right, and be less confident when the models are wrong.

    Moreover, the hypothesis-driven approach reduced over-reliance significantly compared to the standard AI recommendation. Similarly, it also reduced under-reliance compared to AI-explanation-only. Importantly, the positive result for under-reliance using the recommendation-driven approach is not cancelled out by the poor over-reliance result, compared to the hypothesis-driven. The primary aim indicates potential for the use of uncertainty/confidence \cite{Bhatt20uncertainty,le2023explaining} and conformal prediction \cite{Straitouri2023} to direct decision-makers' attention towards a set of hypotheses that it is confident about.

    Using the qualitative analysis, hypothesis-driven helped participants take advantage of the decision support tool's evidence, and also recognise the uncertainty underlying the model. Using the strength of evidence, participants are aware of the uncertainty between multiple hypotheses. Therefore, they made an attempt to gauge the model uncertainty by calibrating the weight of evidence depending on whether the feature is important or not. Also, they could make use of the input feature values and choose the hypothesis that they perceive most likely matches with those values. 

    On the other hand, \textit{recommendation-driven} and \textit{AI-explanation-only} do not support this. We found that in recommendation-driven, people could use feature values to confirm the validity of the decision aid's prediction. However, they were not aware of the uncertainty among different hypotheses. In AI-explanation-only, people often ignore using the evidence and solely focus on using the feature values to make a decision because interpreting the evidence with this approach can be a lot more mentally demanding.

    \subsubsection{Skin cancer diagnosis domain}
    
    Recommendation-driven is preferred by beginners who do not have much experience in the field. It is easier to use and faster to make a decision because they only need to confirm if they want to follow the recommendation or not. However, it can bias the clinicians and make them doubt their own diagnosis.
    
    Hypothesis-driven is more suitable for experienced participants who have knowledge in the field. It allows users to compare and contrast between different hypotheses and check where the AI is looking for each diagnosis. However, it requires users to have a good base of knowledge for all possible hypotheses to be able to evaluate the provided evidence. It can also be overwhelming for users who do not have much experience in the domain. In addition, the evidence provided by the AI can be wrong, which can lead to a wrong decision.

    Moreover, the order of AI in relation to the human decision-maker is also important. We can either put the AI first or the human first, both of which have their own advantages and disadvantages~\cite{janda2019can}. The AI-first approach can be used as a triage tool to reduce the assessment time, but it can face regulatory challenges and over-reliance on AI. The human-first approach can retain the current clinical workflow and use AI as a second opinion, which can increase the sensitivity of the diagnosis. However, it can be time-consuming, with a potential increase in consultation time when there are disagreements between the AI and the human. In our study, recommendation-driven implemented the AI first, while hypothesis-driven could be improved by putting the human opinions first before providing the evidence from the AI.

    \subsection{Laypeople and Skilled Individuals}
    We conducted two human experiments in two different domains: housing price prediction and skin cancer diagnosis. The goal is to have a more diverse set of participants, including both laypeople and skilled individuals.

    In the housing price prediction domain, we found that the hypothesis-driven approach can help participants make better decisions. Note that there were only three possible hypotheses in the first domain, whereas there were seven in the second domain. In case of having too many options, the hypothesis-driven approach can be overwhelming for participants, especially for laypeople or beginners in the field. In the second domain, we found that experienced participants applied their domain knowledge to narrow down possible hypotheses and improve their decision-making overall.

    In both studies, participants agree that hypothesis-driven is more informative. However, the recommendation-driven approach is more straightforward and easier to use. In fact, some skilled individuals prefer the recommendation-driven approach because they already have a good base of knowledge to make the decision. They can use the AI to simply compare and contrast their own thought with the AI's recommendation, leading to a faster decision-making process. In case of having high uncertainty in the decision, hypothesis-driven is preferred by everyone as it provides more information.
    \subsection{Threats to Validity}

    We will identify threats to the validity of the human studies, including both internal and external validity.

    \subsubsection{Housing Price Prediction Domain}
        There is a threat to internal validity due to \textbf{instrumentation}, we ran the experiment on one dataset (Ames Housing), which limits generalisability.  In addition, since we follow the labels from this particular dataset, there is no single ground-truth for the price of a house. The price can vary depending on many factors. Therefore, the experimental participants' tasks are somewhat subjective. Further, this task has only three output classes, so only three hypotheses, and we anticipate the results would be more interesting when we consider more hypotheses. The current form of explanation can be further improved by incorporating contrastive explanations. At present, we present multiple WoEs for different hypotheses, but these cannot be considered as contrastive explanations.

        Regarding the experimental design, there are multiple forms of cognitive forcing, and we tested just one. We could have added another experimental condition where we asked participants to first think about their decision before showing the AI recommendation. This condition may have close performance to the hypothesis-driven approach. Finally, the human experiment is currently conducted with laypeople while experts likely interact with the decision-aiding tool differently from laypeople~\cite{Fogliato22Who}. Another limitation is that we currently use the output labels of the model as the hypotheses, while domain experts may have more and different hypotheses.

    \subsubsection{Skin Cancer Diagnosis Domain}

        \paragraph{Threats to Internal Validity}
        A threat to internal validity for a long human experiment (i.e., required approximately one hour to finish) is the \textbf{maturation}. Participants could become tired over time and lose concentration when doing later tasks. \textbf{Instrumentation} is the next threat that needs to be considered. We use labels from the HAM10000 dataset~\cite{Tschandl18} to measure the performance of participants. But it is important to note that there is no single \textit{ground-truth}. Different experts in the field can still have different opinions on the labels. Therefore, we will need a better approach to measuring the correctness of diagnoses rather than solely relying on these labels. Moreover, our pool of participants is relatively small (14 participants), the number of tasks is limited (16 tasks), trained on a single dataset (HAM10000), and limited number of conditions as we did not consider \textit{no AI} condition in the human experiment. All these factors can affect the significance of the results.

        \paragraph{Threats to External Validity}
        The first threat to external validity is \textbf{sample characteristics}. We focus on using our network to invite experienced diagnosers to participate. This may not generalise to a broader population. Additionally, our participants are not experts. They have background knowledge in the field of skin cancer but are still far from being experts. In our current user study, we categorised them into \textit{experienced} (those who have received clinical training) and \textit{inexperienced} (those who have not received clinical training) participants. Future work should consider recruiting experts in the field, which could provide further insights into how domain expertise might influence reliance on AI.
        
        Secondly, regarding \textbf{ecological validity}, laboratory experiments very often do not reflect the real world. For example, while some images can be straightforward, doctors typically require additional information such as the history of the patient, age, sex, regional images, etc. before making a diagnosis. Relying on a single dermoscopic image can be challenging for participants and may lead to high diagnostic uncertainty. Moreover, there is a concern that while AI can perform extremely well in a controlled environment, their behaviour may differ substantially in clinical settings. That is why none of our experienced participants have used AI to support skin cancer diagnosis in clinical practice. At this stage, there is no formally accredited AI system available, and AI is used solely for research and internal testing purposes.
    
    \subsection{Future Work}

    \subsubsection{Combining Recommendation-driven and Hypothesis-driven}

    The question now is \textit{how should we design the decision-making interface in practice?} The answer is that we can combine the advantages of both interfaces. We can start by allowing users to put forward their hypothesis and evidence first to avoid automation bias~\cite{cabitza2024never}. Following this, they will be shown a ranking of hypotheses based on the level of uncertainty. This allows users to be aware of the AI's recommendation (i.e., the hypothesis with the least uncertainty), as well as the supporting and opposing evidence for all possible hypotheses. Alternatively, we can use conformal prediction~\cite{Straitouri2023} to present multiple hypotheses within a given confidence bound. Uncertainty information can also reduce the number of possible hypotheses when there are too many to choose from, helping users focus on the most likely ones and avoid decision fatigue.

    Moreover, the decision-aid can allow \textit{argumentation} between the decision-maker and the AI. Users can provide their thoughts, including the \textit{evidence} supporting their decisions. The decision-support system will then compare the user's evidence with the AI's evidence, as well as the user's hypothesis with the AI's hypothesis. Users can also modify the AI's evidence based on their domain knowledge. This comparison helps users evaluate their decisions against those of the AI model and contribute to improving the model when they believe it is incorrect. Some recent works have used argumentation-based approaches to better design AI decision support. For example,~\citet{chiang2024enhancing} introduced \textit{devil's advocate} to challenge the AI recommendation or the majority opinion within a group. Alternatively, the devil's advocate can be used to present counter-arguments against the user's decision~\cite{ma2024beyond}.~\citet{ma25towards} proposed the idea of \textit{Deliberative AI}, which allows the human user and the AI to deliberate conflicting evidence and arguments.

    Regarding the presentation of evidence, it is not only about how much a feature contributes to the hypothesis, but also about whether the feature is important or relevant to it. In particular, even if we have strong evidence supporting a hypothesis, if the evidence is not relevant, its weight should be calibrated or even ignored when making the final decision.

    \subsubsection{Human Experiment Design}
    In both domains, we should control the cognitive load across experimental conditions by including a cognitive load measurement such as NASA-TLX scales~\cite{hart1988development} or reaction-time analysis. Since the hypothesis-driven condition can require more cognitive effort than the traditional recommendation-driven condition, it is important to determine whether any performance differences are due to cognitive load or the nature of the approach itself. Future work can consider using these cognitive load measures when evaluating different decision-support approaches. Another aspect to improve the human experiments is assessing whether trust calibration and decision quality improve \textit{over time}. Current experiments only compare the decision-support approaches in short-term settings, without considering long-term effects of repeated exposure.
    

%% file: 6-conclusion.tex
\section{Conclusion}
\label{sec:conclusion}
In this paper, we introduce the implementation of a hypothesis-driven approach by extending the Weight of Evidence framework. The new approach is evaluated in two domains: housing price prediction and skin cancer diagnosis. We found that the hypothesis-driven approach can help participants make better decisions by providing a more complete picture of the underlying decision aid. However, the recommendation-driven approach is more streamlined and easier to use, especially for beginners. Different decision-support approaches have their own pros and cons, but can be combined to provide a better decision-support tool in the future.